\documentclass[10pt,twocolumn,letterpaper]{article}

\usepackage{cvpr}
\usepackage{times}
\usepackage{epsfig}
\usepackage{graphicx}
\usepackage{amsmath}
\usepackage{amssymb}

\usepackage{array}
\usepackage{epstopdf, booktabs}
\usepackage{enumitem}
\usepackage{subfigure}
\usepackage{multirow}
\usepackage{lipsum}
\usepackage[english]{babel}
\usepackage[linesnumbered,ruled]{algorithm2e}
\usepackage{appendix}

\DeclareMathAlphabet{\mathpzc}{OT1}{pzc}{m}{it} 

\newcommand{\superscript}[1]{\ensuremath{^{\textrm{#1}}}}

\newcommand{\modelname}{D\superscript{2}AE}

\newcolumntype{P}[1]{>{\centering\arraybackslash}p{#1}}
\newcolumntype{M}[1]{>{\centering\arraybackslash}m{#1}}

\usepackage[pagebackref=true,breaklinks=true,letterpaper=true,colorlinks,bookmarks=false]{hyperref}

 \cvprfinalcopy 

\def\cvprPaperID{229} 

\ifcvprfinal\fi
\begin{document}
\newcommand*{\affaddr}[1]{#1} 
\newcommand*{\affmark}[1][*]{\textsuperscript{#1}}
\newcommand*{\email}[1]{\texttt{#1}}
\title{Exploring Disentangled Feature Representation Beyond Face Identification}

\author{%
Yu Liu\affmark[1]\thanks{They contributed equally to this work}, Fangyin Wei\affmark[2]\affmark[3]\footnotemark[1], Jing Shao\affmark[2]\footnotemark[1], Lu Sheng\affmark[1], Junjie Yan\affmark[2], Xiaogang Wang\affmark[1]\\
\affaddr{\affmark[1]CUHK-SenseTime Joint Lab,  The Chinese University of Hong Kong}\\
\affaddr{\affmark[2]SenseTime Group Limited}, \affaddr{\affmark[3]Peking University}\\
\email{\{yuliu,lsheng,xgwang\}@ee.cuhk.edu.hk}, \email{weifangyin@pku.edu.cn},\\
\email{\{shaojing,yanjunjie\}@sensetime.com}
}

\maketitle
\thispagestyle{empty}

\begin{abstract}

This paper proposes learning disentangled but complementary face features with minimal supervision by face identification.
Specifically, we construct an identity Distilling and Dispelling Autoencoder (\modelname) framework that adversarially learns the identity-distilled features for identity verification and the identity-dispelled features to fool the verification system.
Thanks to the design of two-stream cues, the learned disentangled features represent not only the identity or attribute but the complete input image.
Comprehensive evaluations further demonstrate that the proposed features not only maintain state-of-the-art identity verification performance on LFW, but also acquire competitive discriminative power for face attribute recognition on CelebA and LFWA.
Moreover, the proposed system is ready to semantically control the face generation/editing based on various identities and attributes in an unsupervised manner.

\end{abstract}

\vspace{-0.3cm}
\section{Introduction}
\label{sec:intro}

Learning distinctive yet universal feature representations has drawn long-lasting attention in the community of face analysis due to its pivotal role in various face-related problems such as face verification and attribute recognition~\cite{sun2014a,schroff2015,liu2017rethinking,cao2013,lu2014,liu2017quality}, as well as generative face modeling and controllable editing~\cite{LuTT17,YanYSL16,LiZZ16e,HuangZLH17, LiZZ16e, LampleZUBDR17}.
Most contemporary methods learn the facial features specific to predefined supervision (\eg identities, attributes)~\cite{ taigman2014a,sun2014c,schroff2015,sun2015,SunLWT15, kumar2009, cao2013, lu2014}, and thus hamper these features to be readily generalized to the feature space for a new task without careful fine-tuning.
For example, without explicit supervision, the learned features are likely not to reflect the connection between two attributes \textit{smile} and \textit{mouth open}, nor to relate identity-relevant attributes like \textit{gender} and \textit{race} closely to identity.
Therefore, learning an almighty feature representation generalizable to any face-related tasks is significant in the field of face analysis and possibly transferable to other fields such as pedestrian analysis.

\begin{figure}
\centering
\includegraphics[width=\linewidth]{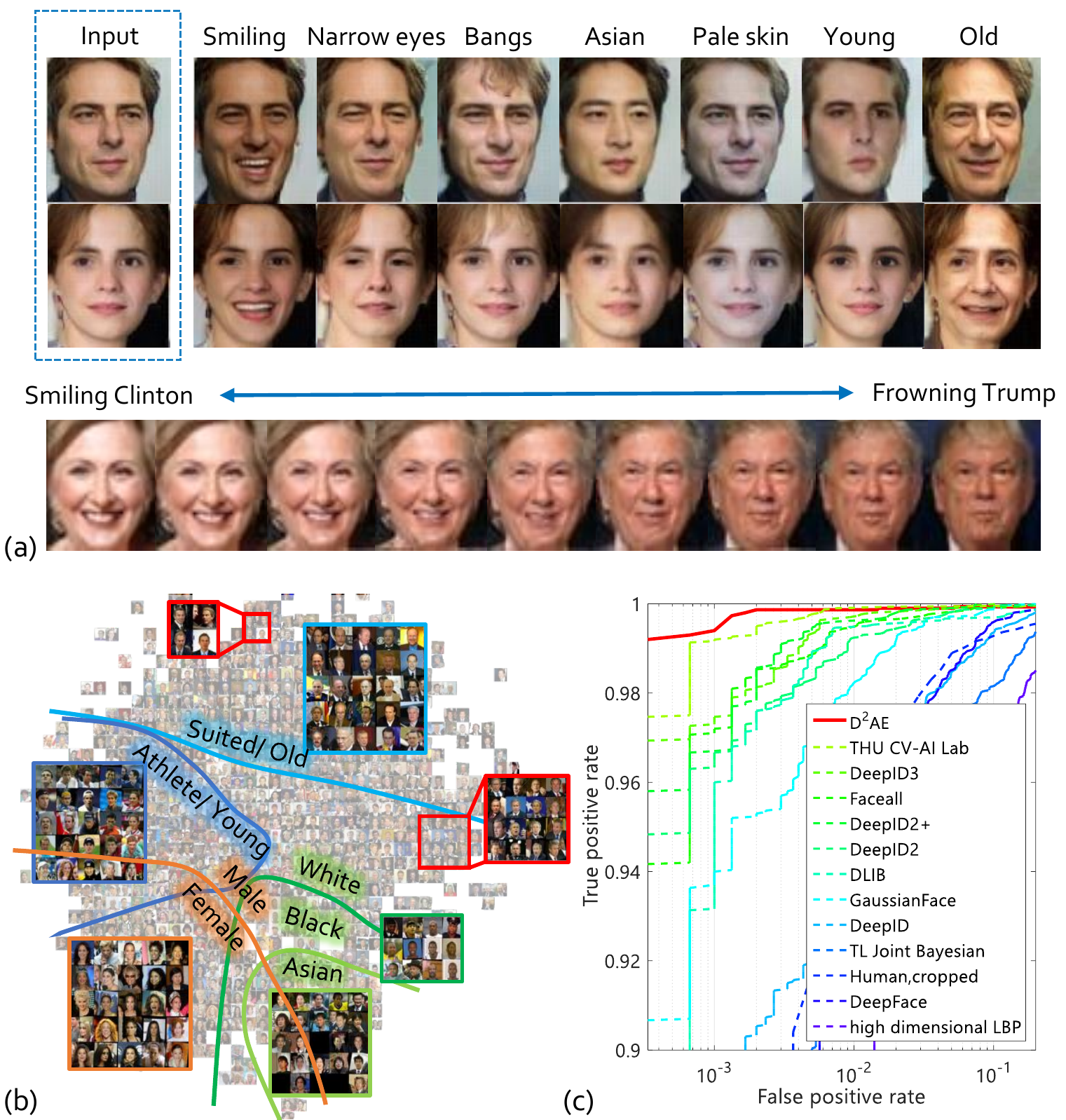}
\caption{Representative face applications based on the learned face feature representations.
(a) Semantic face editing such as identity-preserving attribute modification and identity transfer and interpolation.
(b) The learned face features are trivially separable according to different attributes, visualized by Barnes-Hut \textit{t}-SNE \cite{Maaten2017Visualizing}.
(c) The ROC curve on LFW face verification benchmark. The proposed face feature achieves the accuracy of 99.80\% (single model), which outperforms most state-of-the-art methods without loss of ability in editing identity-related attributes.}
\label{fig:fig_1_intro}
\vspace{-3mm}
\end{figure}

%
Unlike prior arts that applied multi-task supervision~\cite{kumar2009} to extract quasi-universal features that are jointly effective across multiple predefined tasks, in this paper, we propose a novel feature learning framework with a minimal supervision by face identities.
The learned representation not only produces \textit{identity-distilled} features that discriminatively focus on inter-personal differences with identity supervision, but also effectively extracts the hidden \textit{identity-dispelled} features to capture complementary knowledge including intra-personal variances and even background clutters.
Analogous to the adversarial learning paradigm~\cite{goodfellow2014generative,Schmidhuber92c,GaninUAGLLML17}, the identity-dispelled features are fooled to make non-informative judgment over the identities.
We claim that the learned face features own sufficient flexibility to improve face identification and are extensible to model diverse patterns like attributes for different tasks.
Moreover, these features also enable controllable face generation and editing even without tedious training of the control units.
Fig.~\ref{fig:fig_1_intro} illustrates the superiority of the proposed feature representation over the state of the arts in representative applications.

In this study, we wish to highlight four advantages of this innovative feature learning framework:

\noindent (1) \textit{Adversarial Supervision --}
The identity-dispelled features are intactly encoded with the novel adversarial supervision. Distinct from those supervised by additional handcrafted tasks, the proposed scheme is simple yet effectively guarantees better generalization and completeness of the representation with complementary features.

\noindent (2) \textit{Interpretability --}
Our learning scheme provides a comprehensive and decomposable interpretation of the knowledge by adaptively assembling the identity-distilled and identity-dispelled features.
We also find the learned features are compact and smoothly spread in a convex space.
The extracted face features enhence face identity verification and are well prepared for various bypass tasks such as face attribute recognition and semantic face generation/editing.

\noindent (3) \textit{Two-stream End-to-End Framework --}
The proposed framework is end-to-end learned and solely supervised by face identities, distinguished from the conventional methods equipped with alternate adversarial supervision.
By reusing the learned face features, other face-related tasks can be readily plugged in without fine-tuning the network.

\noindent (4) \textit{Discriminative information preserving --}
To be a minor contribution, the performance of face recognition gets improved if the attribute bias against identities occurs in the training set, which is often the case in small datasets.

The aforementioned advantages of the Distilling and Dispelling Autoencoder (\modelname) framework are examined and analyzed through comprehensive ablation studies.
The proposed approach is compared both quantitatively and qualitatively with state of the arts, achieving 1) accuracy of $99.80\%$ on face verification benchmark LFW\cite{lfw}, 2) remarkable performance on attribute classification benchmarks LFWA\cite{liu2015} \& CelebA\cite{liu2015}, and 3) superior capability on various generative tasks such as semantic face editing.
%
%

\vspace{-0.1cm}
\section{Related Work}
\label{sec:related_work}
\vspace{-0.2cm}

\noindent\textbf{Learning Feature Representations.}
With the goal of disentangling distinct but informative factors in the data, representation learning has drawn much attention in the machine learning community~\cite{Bengio2009LearningDA,Bengio2013RepresentationLA}.
It is typically categorized into {generative modeling} and {discriminative modeling}.
Given observations, \emph{Discriminative Models} directly model the conditional probability distribution of the target variables and have accompanied and greatly nourished the rapid progress in classification and regression tasks, such as large-scale facial identity classification \cite{taigman2014a,sun2014a,sun2014c,schroff2015,sun2015,parkhi2015} and attribute classification \cite{ lu2014, chen2013}.
\emph{Generative Models}, as opposed to discriminative models,  learn feature representations by modeling how the data was generated based on the joint distribution of the observed and target variables.
%
%
For example, the autoencoder (AE) framework~\cite{Lecun1987Modeles,Bourlard1988,Hinton1993,hinton2006reducing} proposes that an encoder first extracts features from the data, followed by a decoder that maps from feature space back into input space.
With the ability to automatically encode expressive information from the data space, various AE models \cite{Vincent2008ExtractingAC,Rifai2011ContractiveAE,kingma2013auto} have been developed. 

\noindent\textbf{Combining Discriminative and Generative Models.}
While discriminative models generally perform better, they inherently require supervision, being less flexible than generative models. The pioneering work of GAN \cite{goodfellow2014generative} combines them together, and a large body of literature has been built upon it.
Impressive progress has been made on a variety of tasks, such as image translation \cite{IsolaZZE16}, image editing \cite{ZhuKSE16}, image inpainting \cite{PathakKDDE16, BaoCWLH17}, and texture synthesis \cite{LiW16b,2017arXiv170800315L}.

\noindent\textbf{Disentangled Representation.}
Despite impressive previous progress on improving either visual quality or recognition accuracy, disentangling the feature representation space is still under-explored.
%
%
Some previous works tried to disentangle the representations in tasks such as pose-invariant recognition~\cite{TranYL17, unknown} and identity-preserving image editing~\cite{HuangZLH17, LiZZ16e, LampleZUBDR17}. However, they usually require explicit attribute supervision and encode each attribute as a separate element in the feature vector. 
These methods are limited to representing a fixed number of attributes and need retraining once a new attribute is added.
Makhzani \emph{et al.} ~\cite{MakhzaniSJG15} encode class information into a discrete one-hot vector, with style information following a Gaussian distribution, but its training is likely to be unstable.

Our proposed \modelname{} model overcomes these limitations. With no attribute supervision, the identity-dispelled feature encodes various attributes, to which the identity-distilled feature is invariant.
In contrast, \cite{Yin2017TowardsLF} extracts features that are only pose-invariant, which is a special case of our model. \cite{xiang_pose_invariant_2017} learns a representation that is only invariant to pose and requires multi-source supervisions, while our method learns a representation invariant
to any non-ID attributes and requires no supervision
other than ID.
Moreover, without popular regularization on distribution like VAE~\cite{kingma2013auto}, our learned hidden space is naturally compact and smooth.


\vspace{-0.1cm}
\section{Learning Disentangled Face Features}
\label{sec:face_features}
\vspace{-0.2cm}

In this section, we introduce the identity Distilling and Dispelling Autoencoder (\modelname{}) framework that end-to-end learns disentangled face features with an external supervision signal from face identity.

\begin{figure}
\centering
\includegraphics[width=\linewidth]{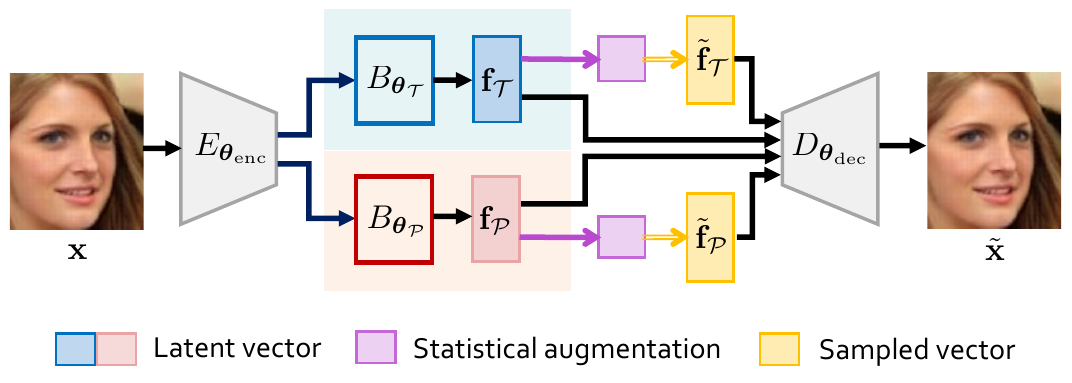}
\caption{The Distilling and Dispelling Autoencoder model.}
\label{fig:fig_2_framework}
\vspace{-3mm}
\end{figure}

Given an input face image $\mathbf{x}$, the identity-distilled feature $\mathbf{f}_\mathcal{T} \in \mathbb{R}^{N_\mathcal{T}}$ and identity-dispelled feature $\mathbf{f}_\mathcal{P} \in \mathbb{R}^{N_\mathcal{P}}$ jointly serve as a complete representation of the face, as illustrated in Fig.~\ref{fig:fig_2_framework}.
The encoding module is composed of  a stack of shared convolutional layers $E_{\boldsymbol\theta_\text{enc}}(\mathbf{x})$, followed by the parallel identity distilling branch $B_{\boldsymbol\theta_\mathcal{T}}$ and identity dispelling branch $B_{\boldsymbol\theta_\mathcal{P}}$.
Face identities supervise the training for $\mathbf{f}_\mathcal{T}$ and also adversarially guide the learning for $\mathbf{f}_\mathcal{P}$.
Finally, a decoding module $D_{\boldsymbol\theta_\text{dec}}(\cdot)$ reconstructs $\tilde{\mathbf{x}}$ from the fused semantic features, so as to encourage the learned face features to encode a full representation of the input image.

\subsection{Identity Distilling Branch}
\label{subsec:identity_distilling_branch}

\begin{figure}
\centering
\includegraphics[width=\linewidth]{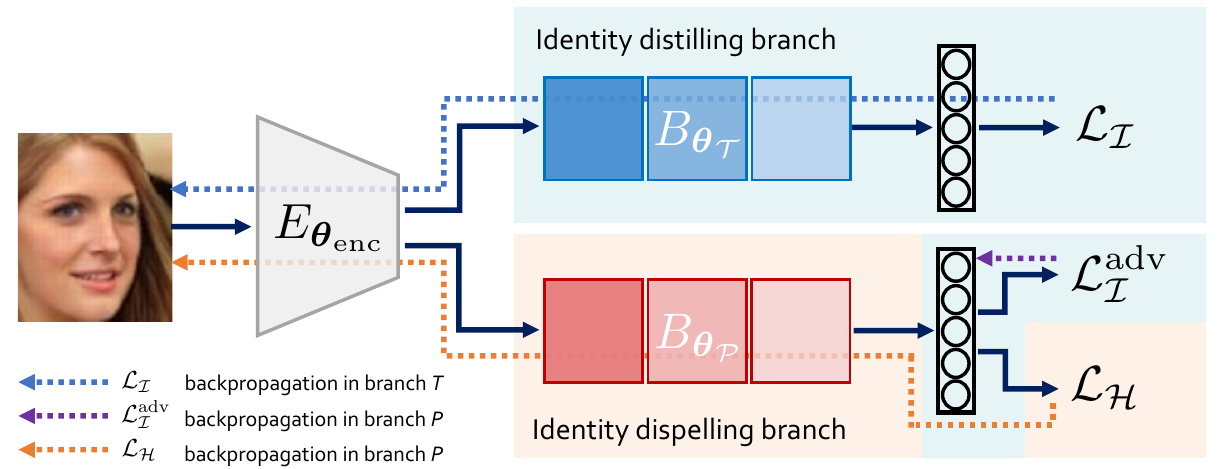}
\caption{The encoding module for extracting disentangled face features.}
\label{fig:fig_3_feature_encoding}
\end{figure}

As visualized in Fig~\ref{fig:fig_3_feature_encoding}, the identity distilling branch \modelname-$\mathcal{T}$ extracts $\mathbf{f}_\mathcal{T}$ by a convolutional subnet $B_{\boldsymbol\theta_\mathcal{T}}$ after $E_{\boldsymbol\theta_\text{enc}}(\mathbf{x})$, written as $\mathbf{f}_\mathcal{T} = B_{\boldsymbol\theta_\mathcal{T}}(E_{\boldsymbol\theta_\text{enc}}(\mathbf{x}))$.
Specifically, $\mathbf{f}_\mathcal{T}$ is non-linearly mapped by \texttt{softmax} function to an $N_\text{ID}$-dimentional identity prediction distribution, which corresponds to the $N_\text{ID}$ identities provided by the applied large-scale training dataset for face identification~\cite{MS1M,liu2015},
\begin{equation}
\vspace{-0.1cm}
\mathbf{y}_\mathcal{T} = \mathtt{softmax}(\mathbf{W}_\mathcal{T}\mathbf{f}_\mathcal{T}+\mathbf{b}_\mathcal{T}).
\vspace{-0.1cm}
\end{equation}
The predicted distribution $\mathbf{y}_\mathcal{T}$ is compared to the ground truth one-hot face labels $\mathbf{g}_\mathcal{I}$ via the cross-entropy loss
\begin{equation}
\vspace{-0.1cm}
\mathcal{L}_\mathcal{I} = \sum_{j=1}^{N_\text{ID}} - \mathbf{g}_\mathcal{I}^j\log\mathbf{y}_\mathcal{T}^j = -\log \mathbf{y}_\mathcal{T}^t,
\vspace{-0.1cm}
\end{equation}
where $t$ indicates the ground truth index.
Please note that the optimization over $\mathcal{L_I}$ only updates the identity distilled branch $B_{\boldsymbol\theta_\mathcal{T}}$ and the shared layers $E_{\boldsymbol\theta_\text{enc}}$.
%

\subsection{Identity Dispelling Branch}
\label{subsec:identity_dispelling_branch}

The identity dispelling branch \modelname-$\mathcal{P}$ suppresses the identity information and tries to encode the complementary facial information.
Similar to the identity distilling branch \modelname-$\mathcal{T}$, it also consists of a subnet $\mathbf{f}_\mathcal{P} = B_{\boldsymbol\theta_\mathcal{P}}(E_{\boldsymbol\theta_\text{enc}}(\mathbf{x}))$ appended with a fully connected layer towards the identity prediction distribution $\mathbf{y}_\mathcal{P} = \mathtt{softmax}(\mathbf{W}_\mathcal{P}\mathbf{f}_\mathcal{P}+\mathbf{b}_\mathcal{P})$.
To enable the complementary feature extraction, we propose an adversarial supervision.

On one hand, we also need to train an identity classifier based on the extracted features $\mathbf{f}_\mathcal{P}$ and supervised by the cross entropy loss $\mathcal{L}_\mathcal{I}^\text{adv} = -\log \mathbf{y}_\mathcal{P}^t$.
The difference between the training of $\mathbf{y}_\mathcal{P}$ and $\mathbf{y}_\mathcal{T}$ is that the gradients of $\mathcal{L}_\mathcal{I}^\text{adv}$ are only back-propagated to the classifier but do not update the preceding layers in $B_{\boldsymbol\theta_\mathcal{P}}$ and $E_{\boldsymbol\theta_\text{enc}}$, analogous to the discriminator in GAN models~\cite{goodfellow2014generative}. 

On the other hand, we need to train the identity dispelling branch to fool the identity classifier, where the so-called ``ground truth'' identity distribution $\mathbf{u}_\mathcal{I}$ is required to be constant over all identities and equal to $\frac{1}{N_\text{ID}}$.
Therefore, it is also equivalent to minimizing the negative entropy of the predicted identity distribution
\begin{equation}
\mathcal{L}_\mathcal{H} = \sum_{j=1}^{N_\text{ID}} \mathbf{u}_\mathcal{I}^j \log \mathbf{y}_\mathcal{P}^j = \frac{1}{N_\text{ID}} \sum_{j=1}^{N_\text{ID}} \log \mathbf{y}_\mathcal{P}^j,
\vspace{-0.1cm}
\end{equation}
where the gradients for $\mathcal{L_H}$ are back-propagated to $B_{\boldsymbol\theta_\mathcal{P}}$ and $E_{\boldsymbol\theta_\text{enc}}$ with the identity classifier fixed.

It is worth mentioning that the proposed adversarial supervision does not introduce degenerated solutions for $\mathbf{f}_\mathcal{P}$ (\eg, non-informative patterns).
However, if we remove the identification loss $\mathcal{L}_\mathcal{I}^\text{adv}$ and allow the gradients in $\mathcal{L}_\mathcal{H}$ to update the identity classifier, few efforts are needed for this branch to deceive $\mathcal{L_H}$ , \eg, by simply changing the identity classifier to produce non-informative outputs.
In this case, there is certainly no guarantee that $\mathbf{f}_\mathcal{P}$ extracts the identity-dispelled features.

The total loss for this branch is the summation of $\mathcal{L}_\mathcal{I}^\text{adv}$ and $\mathcal{L_H}$, and the two features can be learned simultaneously with the proposed feature-level adversarial training, no longer in need of a fragile alternate training process as is required in most GAN models~\cite{goodfellow2014generative}.
%

\subsection{Encoder-Decoder Architecture} 
\label{subsec:auto_encoding_architecture}

While loss functions imposed on identity distilling and dispelling branches encourage a split of the input image represention, there is no guarantee that the combination of $\mathbf{f}_\mathcal{T}$ and $\mathbf{f}_\mathcal{P}$ form a complete encoding of the input image $\mathbf{x}$.
In fact, we can only ensure that $\mathbf{f}_\mathcal{T}$ represents the identity while $\mathbf{f}_\mathcal{P}$ wipes off the identity, but whether the remaining information has been encoded is not clear.
An encoder-decoder architecture is used to further enhance the learned feature embedding by imposing a bijective mapping between an input image and its semantic features.
For simplicity, we apply the $\ell_2$ norm as the reconstruction loss
\begin{equation}
\mathcal{L}_\mathcal{X} = \frac{1}{2}\Vert \mathbf{x} - D_{\boldsymbol\theta_\text{dec}} (\mathbf{f}_\mathcal{T}, \mathbf{f}_\mathcal{P}) \Vert_2^2.
\end{equation}
Since $\mathcal{L}_\mathcal{I}$ encourages $\mathbf{f}_\mathcal{T}$ to distill identity-aware features, the reconstruction loss forces $\mathbf{f}_\mathcal{P}$ to encode all of the remaining identity-irrelevant information to recover the original image.

\subsection{Statistical Augmentation}
\label{subsec:augment}
\vspace{-0.1cm}

To encourage the channel-wise feature distribution in $\mathbf{f}_\mathcal{T}$ and $\mathbf{f}_\mathcal{P}$ to be sufficiently distinctive and concentrated, we may augment the features with Gaussian noises as
\begin{equation}
\tilde{\mathbf{f}}_\iota^i = \mathbf{f}_\iota^i + \varepsilon \boldsymbol\sigma_\iota^i, \forall i \in \{1,\ldots,N_\iota\}~~\text{with}~\varepsilon\sim\mathcal{N}(0,1),
\end{equation}
where $\iota \in \{ \mathcal{T}_{id}, \mathcal{P} \}$ indicates the feature type.
The scale is the standard deviation of each element in $\mathbf{f}_\iota$, which can be efficiently calculated via a strategy similar to batch normalization~\cite{bn}.
When plugging the augmentation operations right after $\mathbf{f}_\mathcal{I}$ and $\mathbf{f}_\mathcal{P}$, the loss functions aforementioned can be straightforwardly modified by the augmented features.

A slight perturbation on $B_{\boldsymbol\theta_\mathcal{T}}$ forces the ID-distilling space to learn larger margins between identities. Furthermore, since the perturbation in each channel is independent, it is useful for channel-decoupling, which is similar to the mechanism of dropout.
%
Therefore, the resultant features are densely concentrated and nearly independent across channels.
%
%
Moreover, it inherently condenses the semantic feature space expanded by $\mathbf{f}_\mathcal{T}$ and $\mathbf{f}_\mathcal{P}$, increasing the network interpretability for any face image.

\subsection{Learning Algorithm}
\vspace{-0.1cm}

Learning the face features involves a single objective that consists of the feature extraction losses ${\mathcal{L}}_\mathcal{I}$, ${\mathcal{L}}_\mathcal{I}^\text{adv}$ and ${\mathcal{L}}_\mathcal{H}$, as well as the reconstructed loss ${\mathcal{L}}_\mathcal{X}$.
Moreover, when statistically augmented by $\tilde{\mathbf{f}}_\mathcal{T}$ and $\tilde{\mathbf{f}}_\mathcal{P}$, we also incorporate the objective with the augmented reconstruction loss $\tilde{\mathcal{L}}_\mathcal{X}$.
The final objective is a weighted combination: 
\begin{equation}
\mathcal{L} = \lambda_\mathcal{T} {\mathcal{L}}_\mathcal{I} + \lambda_\mathcal{P} \left( {\mathcal{L}}_\mathcal{I}^\text{adv} + {\mathcal{L}}_\mathcal{H}\right) + \lambda_\mathcal{X} \left( \tilde{\mathcal{L}}_\mathcal{X} + \mathcal{L}_\mathcal{X}  \right).
\end{equation}
\vspace{-0.4cm}

We apply the stochastic gradient descent solver to minimize the above objective and update the network parameters.
%
%
As depicted in Fig.~\ref{fig:fig_3_feature_encoding}, the dotted blue line and the dotted orange line present the back-propagation routines for  $\mathcal{L_I}$ and $\mathcal{L_H}$, respectively, and the purple line demonstrates the simultaneous back-propagation path for $\mathcal{L}_\mathcal{I}^\text{adv}$.
Similarly, the gradient updates for the encoder-decoder network parameters are back-propagated through the whole autoencoder except the identity classifiers for both branches.

%

\section{Experimental Setting} 
\label{sec:experimental_setting}

\subsection{Datasets and Preprocessing}
\label{sub:datasets_and_preprocessing}
\vspace{-0.1cm}

\noindent\textbf{Datasets.} The proposed \modelname{} model is trained on the MS-Celeb-1M dataset~\cite{MS1M}, which is currently the largest face recognition dataset.
For purpose of assessing its generalization ability, the trained model is evaluated on the LFW dataset~\cite{liu2015}, and the overlapped images both in the MS-Celeb-1M and LFW datasets are manually pruned from the MS-Celeb-1M dataset.
Therefore, $4$M checked images with $80$K identities in the MS-Celeb-1M dataset are used for training and validation, with a split ratio of $9:1$.

\noindent\textbf{Preprocessing.} Faces in the images are detected and aligned by RSA~\cite{RSA}.
Face patches are first cropped so that the interpupillary distance is equal to $35\%$ of the patch width, and then they are resized to $235\times235$.

\subsection{Detailed Implementation}
\label{sub:detailed_implementation}
\vspace{-0.1cm}

The proposed \modelname{} model
consists of an encoding module $E_{\boldsymbol\theta_\text{enc}}$, two parallel subnets $B_{\boldsymbol\theta_\mathcal{T}}$ and $B_{\boldsymbol\theta_\mathcal{P}}$ to decompose the face features, and a decoding module $D_{\boldsymbol\theta_\text{dec}}$.

\noindent\textbf{Encoding Module $E_{\boldsymbol\theta_\text{enc}}$.} We use Inception-ResNet\cite{szegedy2017inception} as the backbone of $E_{\boldsymbol\theta_\text{enc}}$. The input size is modified to $235\times235$ and the final \texttt{AvePool} layer is replaced by $B_{\boldsymbol\theta_\mathcal{T}}/B_{\boldsymbol\theta_\mathcal{P}}$.
%

\noindent\textbf{Subnets $B_{\boldsymbol\theta_\mathcal{T}}/B_{\boldsymbol\theta_\mathcal{P}}$.}
%
%
Each subnet has $3$ \texttt{conv} layers, one global \texttt{AvePool} and one \texttt{FC} layer.
These branches extract two $256$ dimensional feature representations for $\mathbf{f}_\mathcal{T}$ and $\mathbf{f}_\mathcal{P}$.

\noindent\textbf{Decoding Module $D_{\boldsymbol\theta_\text{dec}}$.}
$D_{\boldsymbol\theta_\text{dec}}$ decodes the concatenation $\{\mathbf{f}_\mathcal{T}, \mathbf{f}_\mathcal{P}\}$ into a face image with the same size as the input image.
The concatenated feature vectors are firstly passed into an $\mathtt{FC}$ layer to increase the feature dimension and then reshaped to squared feature maps, which are fed into $20$ \texttt{conv} layers interlaced with $6$ \texttt{upsampling} layers to obtain the output image.

\noindent\textbf{Model Training.}
The whole network is trained in an end-to-end manner with all of the supervisory signals simultaneously added to the system.
The batch size of the input images is {192}, distributed on 16 NVIDIA Titan X GPUs.
The base learning rate is set to $0.01$ and is declined by $0.1$ every $10$ epochs.
It takes around $31$ epochs in total for the training to converge.
The weights in the training objective is set as {$\lambda_\mathcal{T} = 1$} for $\mathcal{L}_\mathcal{I}$, {$\lambda_\mathcal{P} = 0.1$} for $\mathcal{L}_\mathcal{I}^\text{adv}$ and $\mathcal{L}_\mathcal{H}$, and { $\lambda_\mathcal{X}=1.81\times10^{-5}$} for the $\mathcal{L}_\mathcal{X}$ and $\tilde{\mathcal{L}}_\mathcal{X}$ in the encoder-decoder architecture.

\subsection{Model Evaluation}
\label{sub:model_evaluation}
\vspace{-0.1cm}

We select three representative face-related applications to demonstrate the effectiveness of the proposed face features.
They share the same feature extraction pipeline that concatenates the face features from the proposed identity distilling and dispelling branches $\mathbf{f}_\mathcal{C}^\top = [\mathbf{f}_\mathcal{T}^\top, \mathbf{f}_\mathcal{P}^\top]$.
%

\noindent\textbf{Face Identification.}
We select the LFW dataset as the test bed for face identification, following the standard evaluation protocols with two popular metrics: accuracy and TPR@$0.001$FPR\footnote{We take TPR for short in the following experiments.}.
The identity similarity is calculated by the cosine distance between two feature vectors.

\noindent\textbf{Face Attribute Recognition.}
We further validate the discriminative power of the proposed face features on face attribute recognition over the CelebA~\cite{liu2015} and LFWA~\cite{liu2015} datasets.
Each image in these datasets is annotated with $N_\text{att} = 40$ face attributes.
The performance is evaluated by the metric of accuracy, as suggested by Liu~\etal~\cite{liu2015}.
Since our model does not receive the attribute supervision, we extract the combined features $\mathbf{f}_\mathcal{C}$ and then train a linear SVM supervised by the labeled attributes in these datasets.

\noindent\textbf{Face Editing.}
We also show the superiority of the proposed model in identity-preserving attribute editing and attribute-preserving identity exchanging.
\footnote{To prove the robustness and consistency of our model, identities of visualized results are re-used for multiple times in the main paper.}
By editing the semantic face features within the valid range of the feature space, we can observe rich semantic variations in the decoded image.
(1) The identity-preserving attribute editing modifies $\mathbf{f}_\mathcal{P}$ by adding an incremental vector along the max-margin direction $\mathbf{w}_n$ of an attribute according to the trained linear SVM classifier for face attribute recognition.
Thus the modified feature is $\mathbf{f}_\mathcal{P}^* = \mathbf{f}_\mathcal{P} + \alpha_n \mathbf{w}_n$, where $\alpha_n \mathbf{w}_n$ ranges within the confidence interval controlled by the learned standard deviation depicted in Sec.~\ref{subsec:augment} for a reasonable modification of the input image.
To support editing multiple attributes, $\mathbf{f}_\mathcal{P}^*$ can be extended to $\mathbf{f}_\mathcal{P}^* = \mathbf{f}_\mathcal{P} + \sum_{n=1}^{N_\text{att}} \alpha_n \mathbf{w}_n$, where $\boldsymbol\alpha$ is constrained in a similar fashion.
(2) The attribute-preserving identity exchanging replaces $\mathbf{f}_\mathcal{T}^A$ from the image of one identity $A$ with $\mathbf{f}_\mathcal{T}^B$ from another identity $B$, while keeping $\mathbf{f}_\mathcal{P}^A$ unchanged.
Or more generally, the identity can be smoothly varied along the identity manifold, such as $\mathbf{f}_\mathcal{T}^* = \beta \mathbf{f}_\mathcal{T}^A + (1-\beta)\mathbf{f}_\mathcal{T}^B, \forall \beta \in [0, 1]$.
The generated face image has the target identity with the rest semantic information and background remaining the same.
%

\begin{table}[t]
\center
\footnotesize{
\begin{tabular}{M{2.4cm}|M{0.9cm}M{0.9cm}M{0.9cm}M{0.9cm}}
\hline
\multirow{2}{*}{\textbf{Branch}} & \multicolumn{2}{c}{\textbf{Identity}} & \multicolumn{2}{c}{\textbf{Attribute}} \\
\cline{2-5}
 & \textbf{Acc} & \textbf{TPR} & \textbf{Acc} & \textbf{\#drop}\\
\hline
\modelname-$\mathcal{T}$ & \underline{99.78} & \textbf{99.63} & 79.78 & --\\
\modelname-$\mathcal{P}$ & 64.13 & 5.3 & \underline{81.99} & --\\
\hline
\modelname-$\mathcal{P}$ w/o $\mathcal{L}_\mathcal{H}$ & 71.2 & 8.67 & 80.47 & 36/40 \\
\modelname-$\mathcal{P}$ w/o $\mathcal{L}_\mathcal{I}^\text{adv}$ & 67.13 & 5.63 & 78.32 & 36/40 \\
\hline
\modelname & \textbf{99.80} & 99.40 & \textbf{83.16} & --\\
\hline
\end{tabular}
}
\vspace{+1mm}
\caption{Evaluation of the \modelname{} model on identity verification and attribute recognition, comparing different combinations of branches and losses. The last column shows the number of attributes (out of the total number of $40$) that suffer a performance drop compared to \modelname-$\mathcal{P}$ with complete losses. Bold font marks the best result in each column.}
\label{tb:ablation_branch}
\vspace{-3mm}
\end{table}


\vspace{-0.1cm}
\section{Ablation Study}
\label{sec:ablation_study}
\vspace{-0.2cm}

A unique advantage of the \modelname{} model is its capability of learning complete and disentangled features from the input image, \ie, the identity-distilled feature and identity-dispelled feature.
Successful extraction of the expected features is guaranteed by several pivotal components, \ie, two complementary branches for information selectivity, adversarial supervision for identity dispelling and statistical augmentation for a compact hidden space.
In this section, we will validate their effectiveness by ablation studies, where the LFW(A) face dataset is employed for evaluation.


\begin{figure}
\centering
\subfigure[$\mathbf{f}_\mathcal{T}$ by \modelname-$\mathcal{T}$]{\includegraphics[width=0.45\linewidth]{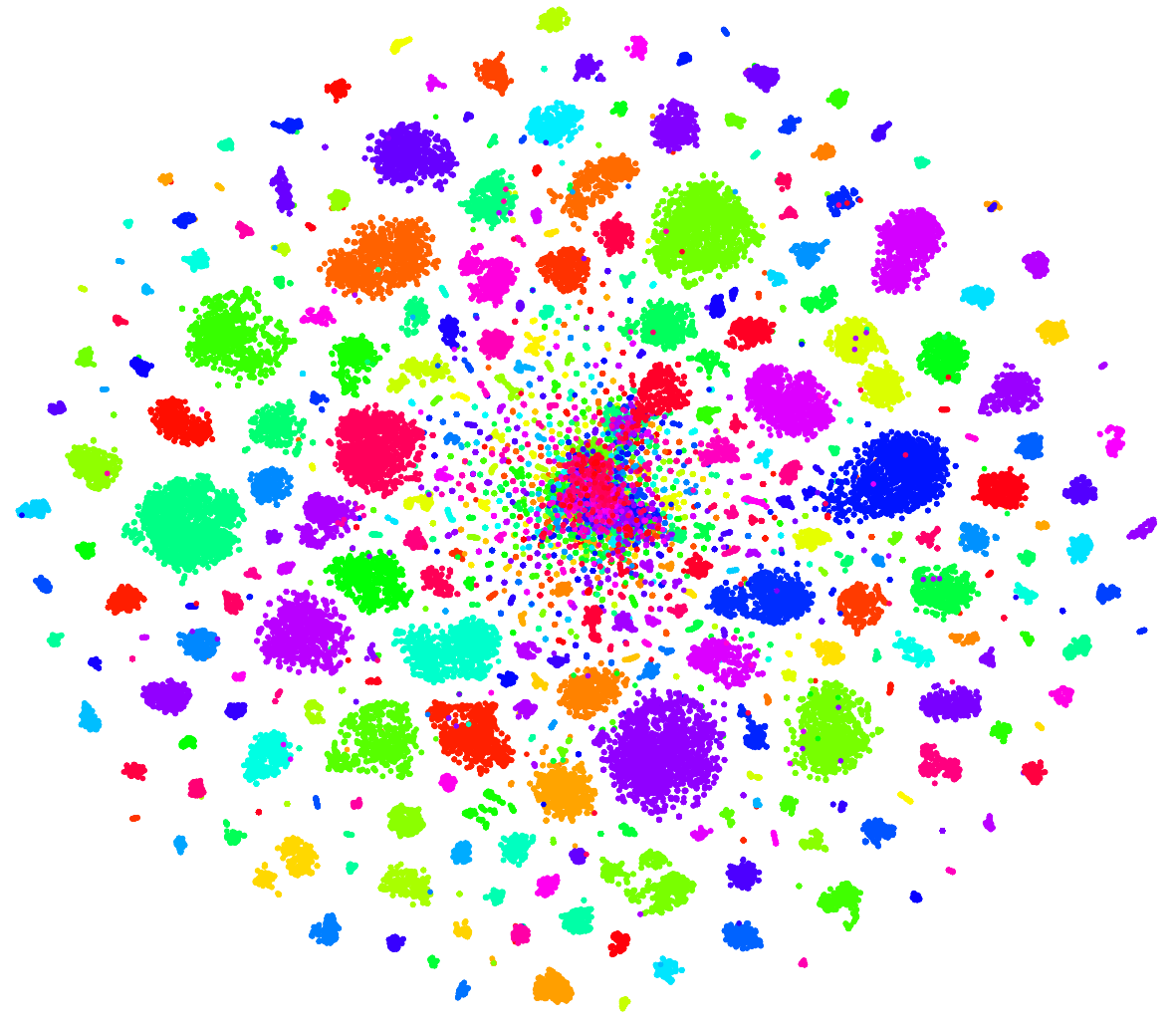}}
\subfigure[$\mathbf{f}_\mathcal{P}$ by \modelname-$\mathcal{P}$]{\includegraphics[width=0.45\linewidth]{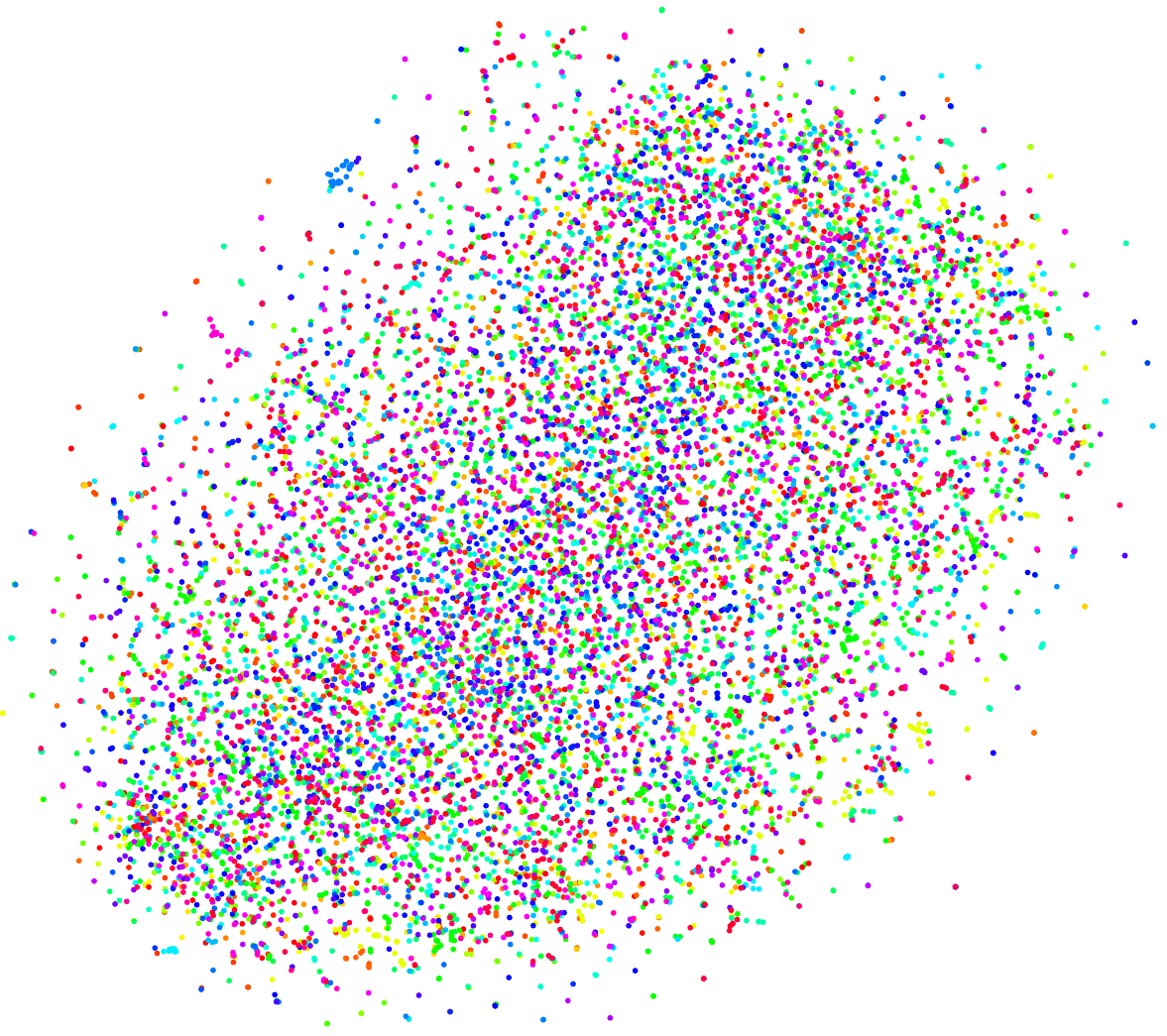}}
\caption{Barnes-Hut \textit{t}-SNE \cite{Maaten2017Visualizing} visualization of the features extracted by two branches (a) \modelname-$\mathcal{T}$ and (b) \modelname-$\mathcal{P}$ on LFW. The colors indicate different identities. Best viewed in color.}
\vspace{-0.2cm}
\label{fig:ablation_branch_id_tsne}
\end{figure}

\subsection{Branch Selectivity}
\label{subsec:ablation_branch}
\vspace{-0.1cm}

%
We find that the identity distilling branch \modelname-$\mathcal{T}$ and the identity dispelling branch \modelname-$\mathcal{P}$ indeed have distinctive capacities in representing different features.

\vspace{+0.5mm}
\noindent\textbf{Identity-distilled Feature $\mathbf{f}_\mathcal{T}$.}
Comparing TPR of identity verification in Table~\ref{tb:ablation_branch}, $\mathbf{f}_\mathcal{T}$ from \modelname-$\mathcal{T}$ is significantly superior to $\mathbf{f}_\mathcal{P}$ from \modelname-$\mathcal{P}$.
The extremely low value in TPR for \modelname-$\mathcal{P}$ indicates that this branch has expelled most of the identity-related information from the input image.
To further demonstrate their discrepancy on discriminative capability, we visualize the high-level features generated by these branches based on Barnes-Hut $t$-SNE \cite{Maaten2017Visualizing}.
As shown in Fig.~\ref{fig:ablation_branch_id_tsne} (a), \modelname-$\mathcal{T}$ generates a set of densely clustered features for each identity with distinct boundaries between features from different identities.
Moreover, \modelname-$\mathcal{T}$ has almost the same identity verification result as that of the combined features (named as \modelname{} in Table~\ref{tb:ablation_branch}) and even outperforms the latter by the TPR metric.
Not surprisingly, it also proves that the features by \modelname-$\mathcal{T}$ have an extraordinary ability to represent identity-aware information.

\begin{figure}[t]
\centering
\includegraphics[width=0.95\linewidth]{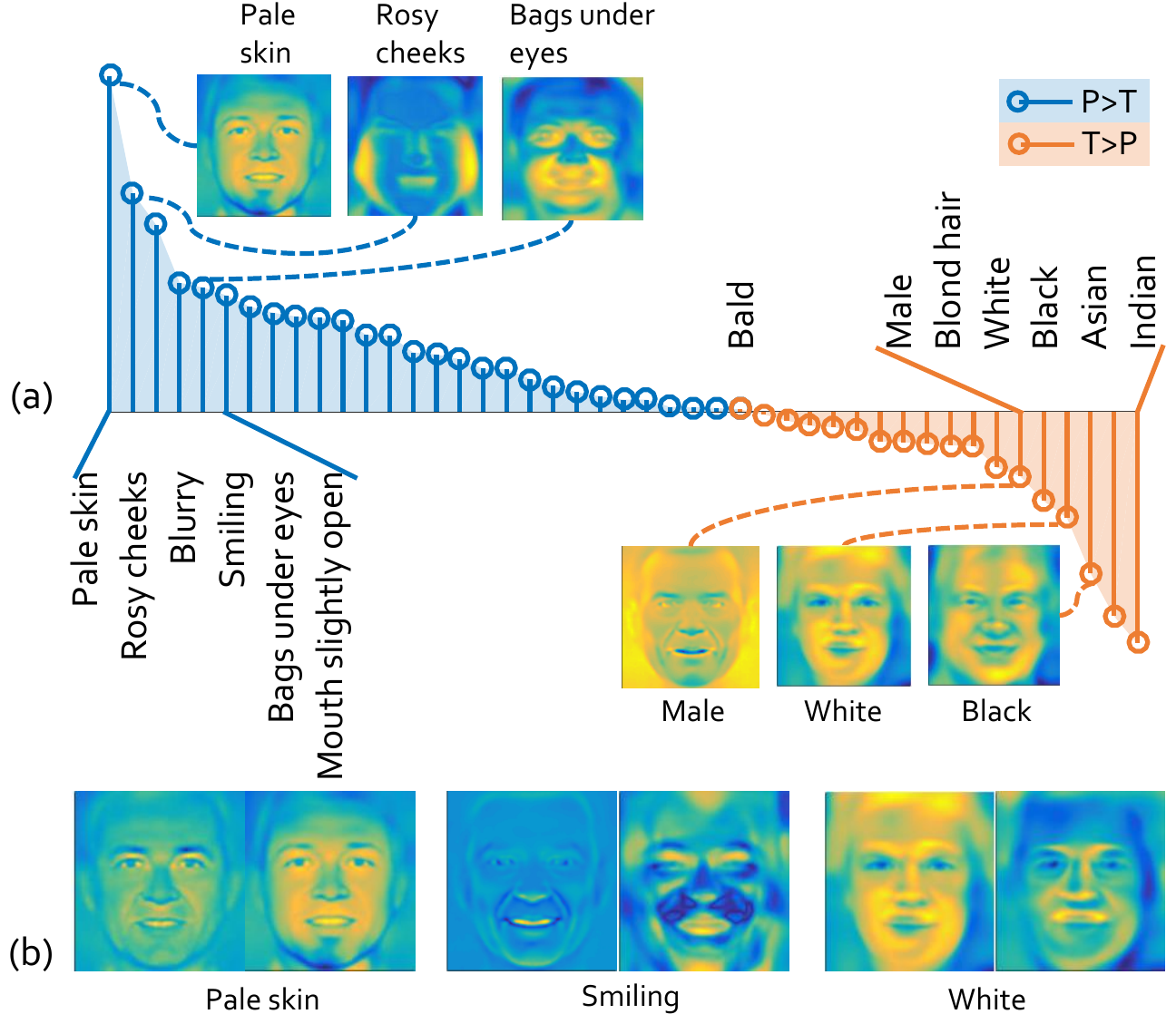}
\caption{(a) Performance comparison on attribute recognition with features extracted from either \modelname-$\mathcal{T}$ or \modelname-$\mathcal{P}$.
	The magnitude of each bin illustrates the difference between the recognition accuracy of \modelname-$\mathcal{T}$ and that of \modelname-$\mathcal{P}$.
	Blue bins indicate attributes where features from \modelname-$\mathcal{P}$ excel, and red bins show attributes where features from \modelname-$\mathcal{T}$ win.
	(b) The residual maps correspond to three representative attributes. Images on the left of each pair are generated by \modelname-$\mathcal{T}$ and the right ones are by \modelname-$\mathcal{P}$.
}
\label{fig:attribute_xaea_xaeb}
\vspace{-0.2cm}
\end{figure}

\vspace{+0.5mm}
\noindent\textbf{Identity-dispelled Feature $\mathbf{f}_\mathcal{P}$.}
In contrast to its poor ability of extracting identity-aware features, \modelname-$\mathcal{P}$ presents its superiority in face attribute recognition over \modelname-$\mathcal{T}$ , as shown in Table~\ref{tb:ablation_branch}.
Interestingly, the features learned by \modelname-$\mathcal{T}$ also present certain discriminative ability to recognize some attributes.
As shown in Fig.~\ref{fig:attribute_xaea_xaeb}, the feature $\mathbf{f}_\mathcal{P}$ outperforms $\mathbf{f}_\mathcal{T}$ on $27$ attributes in total.
For most common attributes that are independent from identity such as \textit{pale skin} and \textit{smile}, the identity-dispelled feature exhibits more discriminative potential.
However, other identity-aware attributes including genders (\eg, \textit{male}) and races (\eg, \textit{Indian} and \textit{Asian}) tend to be better recognized through the identity-distilled feature.
Besides, attributes on the borderline (\eg, \textit{bald}) with similar performance shared by $\mathbf{f}_\mathcal{T}$ and $\mathbf{f}_\mathcal{P}$, are mostly vaguely defined between identity-related and identity-irrelevant attributes.

To visualize the discriminative response of attribute onto the image space, we synthesize a set of residual images responsive to each attribute against the mean image from the LFW dataset.
We synthesize the attribute-augmented face image by first adding a unit vector $\mathbf{w}_n, n\in\{1,\ldots, N_\text{att}\}$ to the mean feature $\bar{\mathbf{f}}_\mathcal{T}$ (or $\bar{\mathbf{f}}_\mathcal{P}$) and then decoding the combined feature to a face image.
%
%
The residual images are attribute-augmented face images subtracted by the mean image.
According to the results shown in Fig.~\ref{fig:attribute_xaea_xaeb} (a), residual maps with respect to \modelname-$\mathcal{P}$ for identity-irrelevant attributes usually display high responses at local semantic regions, such as the facial skins for \textit{pale skin} and cheeks for \textit{rosy cheeks}.
In contrast, the residual maps with respect to \modelname-$\mathcal{T}$ for identity-related attributes usually have holistic responses, which are distributed throughout the whole image, \eg, the maps for \textit{gender} and \textit{race}.
Comparing the residual maps with respect to each branch, $\mathbf{f}_\mathcal{T}$ tends to be more responsive to identity-aware attributes while $\mathbf{f}_\mathcal{P}$ displays stronger responses to identity-irrelevant attributes, as presented in Fig.~\ref{fig:attribute_xaea_xaeb} (b).

\begin{figure}
\centering
\includegraphics[width=\linewidth]{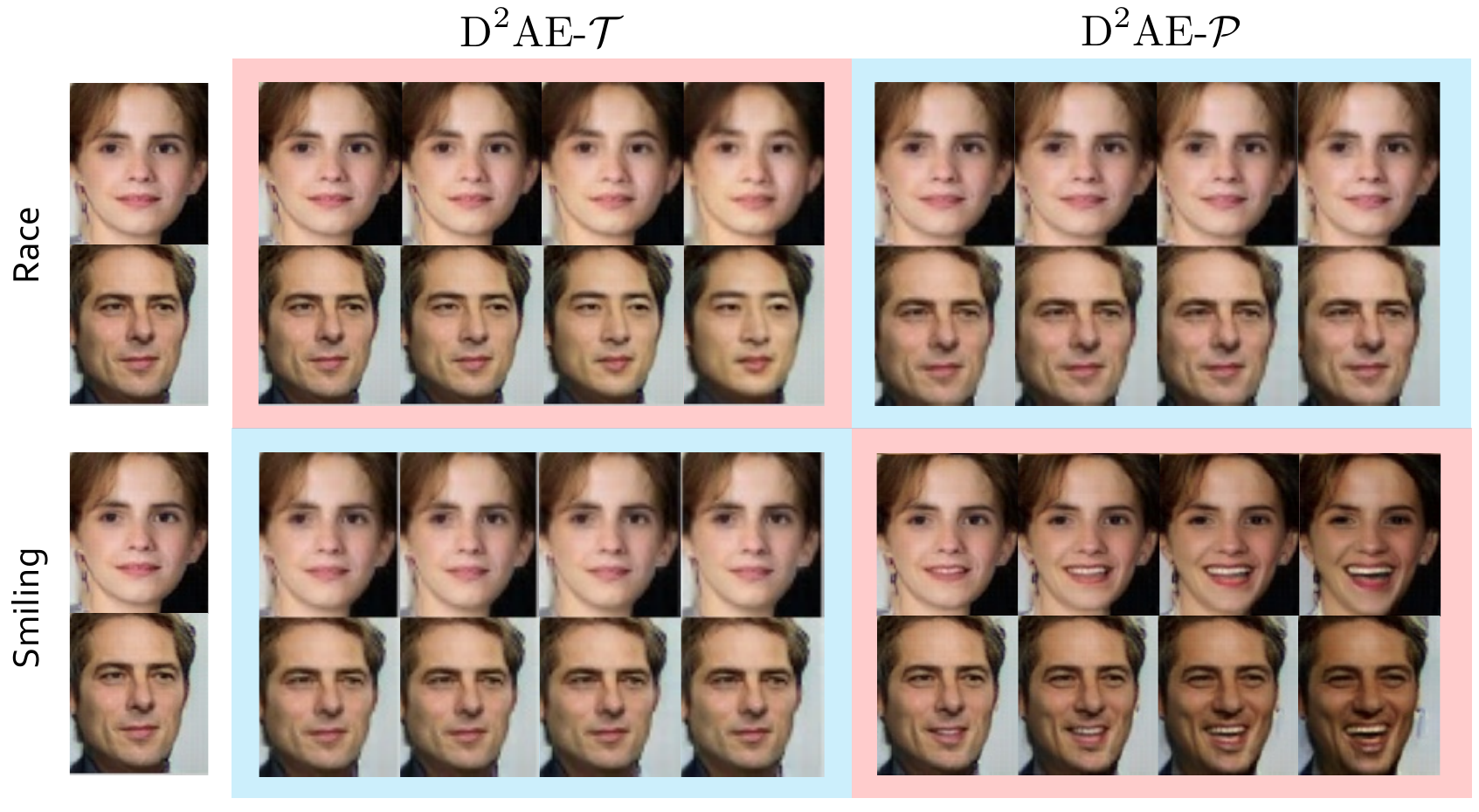}
\vspace{-0.5cm}
\caption{Modifying features extracted from two branches for face attribute editing. ID-related and ID-irrelevant attributes are extracted through \modelname-$\mathcal{T}$ and \modelname-$\mathcal{P}$, respectively.  }
\label{fig:face_attribute_editing_branch_comparison}
\vspace{-0.2cm}
\end{figure}

In addition to quantitative comparison, in Fig.~\ref{fig:face_attribute_editing_branch_comparison} we also provide qualitative results of face attribute editing by modifying features extracted from \modelname-$\mathcal{T}$ and \modelname-$\mathcal{P}$.
Modifying $\mathbf{f}_\mathcal{T}$ has minimal influence on face variation when editing identity-irrelevant attributes like \textit{smiling}, but it effectively controls the face warping to a different \textit{race}.
Conversely, modification on $\mathbf{f}_\mathcal{P}$ can hardly change the race of a face, but it continuously transforms a face from a neutral expression to \textit{smiling}.

\vspace{-0.1cm}
\subsection{Loss Functionality}
\label{subsec:ablation_loss}
\vspace{-0.2cm}


The adversarial training in the identity dispelling branch is a distinctive feature of the \modelname{} network.
We examine the loss terms in \modelname-$\mathcal{P}$ to demonstrate their necessities for the effective training of identity-dispelled features.

\vspace{+0.5mm}
\noindent\textbf{Identity Confusion Loss $\mathcal{L}_\mathcal{H}$.}
Removing the adversarial loss $\mathcal{L}_\mathcal{H}$ in \modelname-$\mathcal{P}$ causes a failure of $\mathbf{f}_\mathcal{P}$ to completely dispel identity-related attributes.
As a consequence, it performs better on identity verification than the model with combined losses, but its performance on attribute recognition is slightly degraded, as presented in Table~\ref{tb:ablation_branch}.
$36$ out of $40$ attributes experience drop of recognition accuracy.
In contrast, both accuracy and TPR metrics for identity verification obtain remarkable gains.
In fact, the identity classification loss cannot effectively constrain $\mathbf{f}_\mathcal{P}$ as it only has an impact on the identity classifier during gradient update, thus there is no guarantee that the resultant $\mathbf{f}_\mathcal{P}$ is independent from the identity as expected.

\vspace{+0.5mm}
\noindent\textbf{Identity Classification Loss $\mathcal{L}_\mathcal{I}^\text{adv}$.}
Removing the identity classification loss $\mathcal{L}_\mathcal{I}^\text{adv}$, we only regularize $\mathbf{f}_\mathcal{P}$ to fool the identity verification system based on $\mathcal{L}_\mathcal{H}$ which updates its identity classifier.
Thus its identity dispelling ability, \ie, the ability of pruning identity information from $\mathbf{f}_\mathcal{P}$, is weaker than the combined losses for lack of explicit identity supervision on the identity classifier.
According to Table~\ref{tb:ablation_branch}, without $\mathcal{L}_\mathcal{I}^\text{adv}$, the performance of $\mathbf{f}_\mathcal{P}$ on identity verification is slightly improved, but its performance on attribute recognition is degraded with drops happening in $36$ over $40$ attributes.
As $\mathcal{L}_\mathcal{H}$ explicitly confuses $\mathbf{f}_\mathcal{P}$ about the identity, it renders poorer identity verification than that trained by $\mathcal{L}_\mathcal{I}^\text{adv}$.
Moreover, with a weaker ability to extract the information complementary to identity, $\mathcal{L}_\mathcal{H}$ also produces inferior attribute recognition results than that by $\mathcal{L}_\mathcal{I}^\text{adv}$.

%
\begin{figure}
\centering
\includegraphics[width=\linewidth]{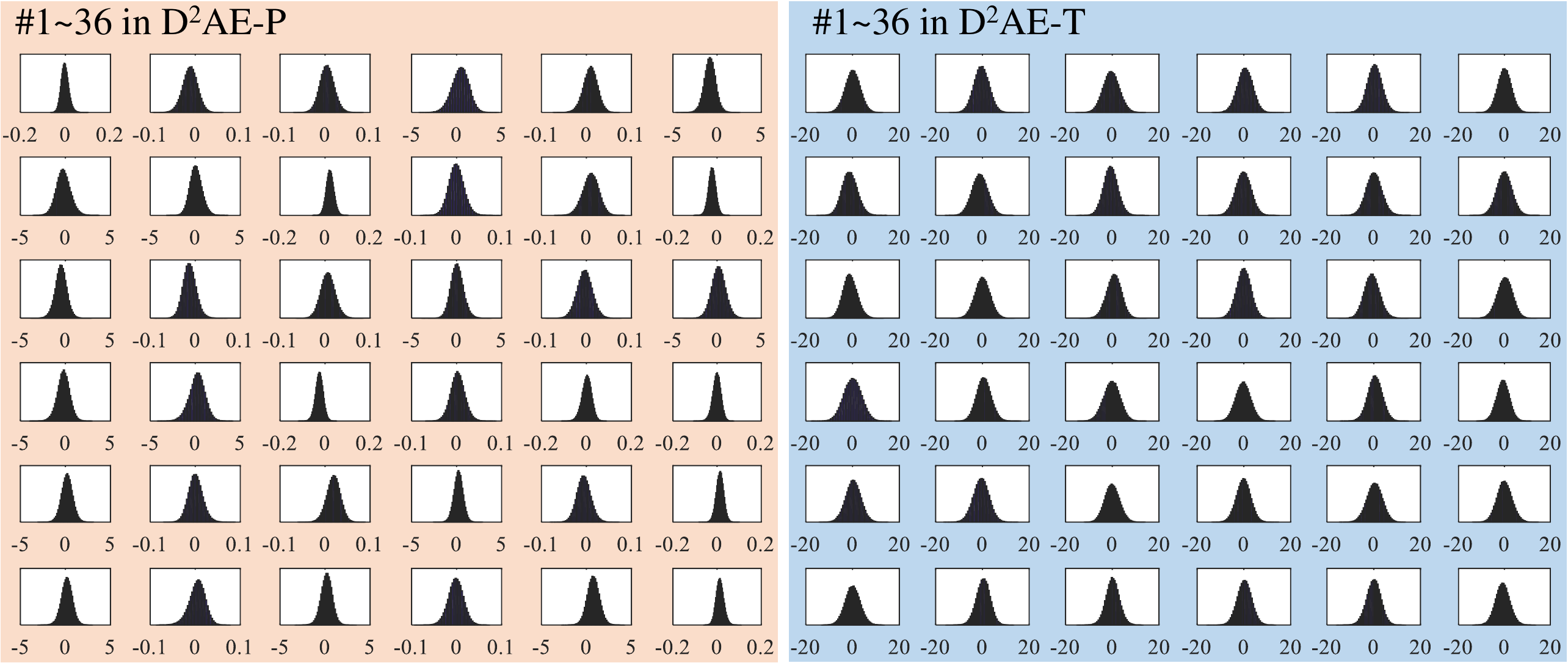}
\caption{The distributions of the first 36 channels in features generated by \modelname-$\mathcal{P}$ (left) and \modelname-$\mathcal{T}$ (right). All the variables follow the Gaussian distribution.}
\label{fig:xiaosunjian}
\vspace{-0.1cm}
\end{figure}

\begin{figure}
\centering
\includegraphics[width=0.95\linewidth]{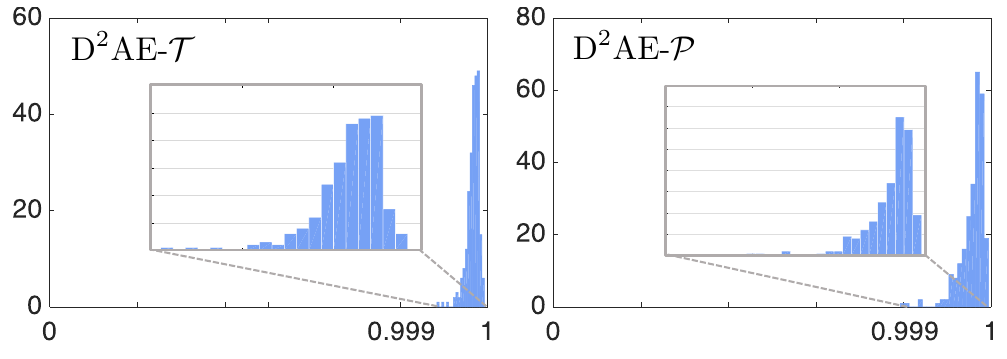}
\vspace{-0.1cm}
\caption{$adj-R^2$ scores of the channel-wise statistics of the features generated by \modelname-$\mathcal{T}$ (left) and \modelname-$\mathcal{P}$ (right).}
\label{fig:feature_gaussian}
\end{figure}

\subsection{Augmentation Necessity for Convex Space}
\label{subsec:ablation_augment}

The proposed statistical augmentation encourages the learned face features to be distinctive and densely Gaussian clustered in each channel as in Fig.~\ref{fig:xiaosunjian}. 
For quantitative evaluation, in Fig.~\ref{fig:feature_gaussian}, we plot two histograms to verify the required statistical property of $\mathbf{f}_\mathcal{T}$ and $\mathbf{f}_\mathcal{P}$ on the LFW dataset.
These histograms are used to mimic the adjusted R-square ($adj-R^2$) score distribution for the channel-wise statistics of the features, where $adj-R^2$ is to measure how much the statistics look like a Gaussian distribution (a higher score is more alike).
Obviously, both features are nearly Gaussian and almost all the channels have $adj-R^2$ scores higher than $0.99$.
They prove that the learned feature spaces for $\mathbf{f}_\mathcal{T}$ and $\mathbf{f}_\mathcal{P}$ are Gaussian and convex, whilst the features in these spaces are densely spread.

\begin{figure}
\centering
\includegraphics[width=\linewidth]{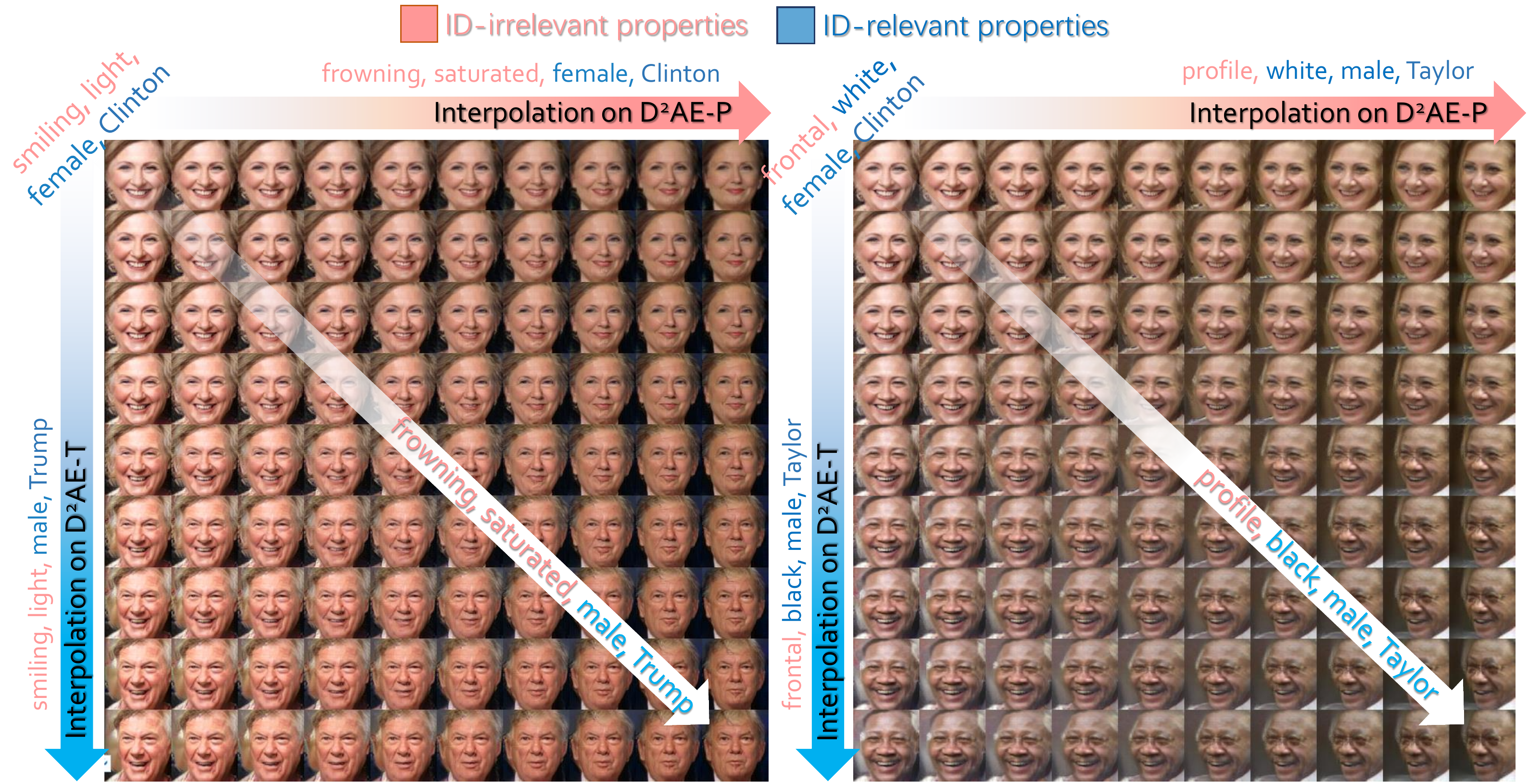}
\caption{Images generated by dense interpolation on space expanded by \modelname-$\mathcal{T}$ and \modelname-$\mathcal{P}$. \modelname{} learns a convex hyper space. \modelname-$\mathcal{T}$ controls identity and all identity-related attributes, such as \textit{gender} and \textit{race}. \modelname-$\mathcal{P}$ controls all the other attributes like \textit{smile} and \textit{frontal}. Best viewed in color and zoomed in.}
\label{fig:clinton2trump}
\vspace{-0.2cm}
\end{figure}

We also find that the learned feature space is compact and convex by densely interpolating two identities with different attributes.
In Fig.~\ref{fig:clinton2trump}, the interpolated face images change smoothly along the identity and attribute axes.

\section{Performance Comparison}
\label{sec:comparison}

We also quantitatively and qualitatively compare the proposed \modelname{} model with state-of-the-art approaches on the face-related tasks as mentioned above.

\begin{figure*}
\centering
\includegraphics[width=0.88\linewidth]{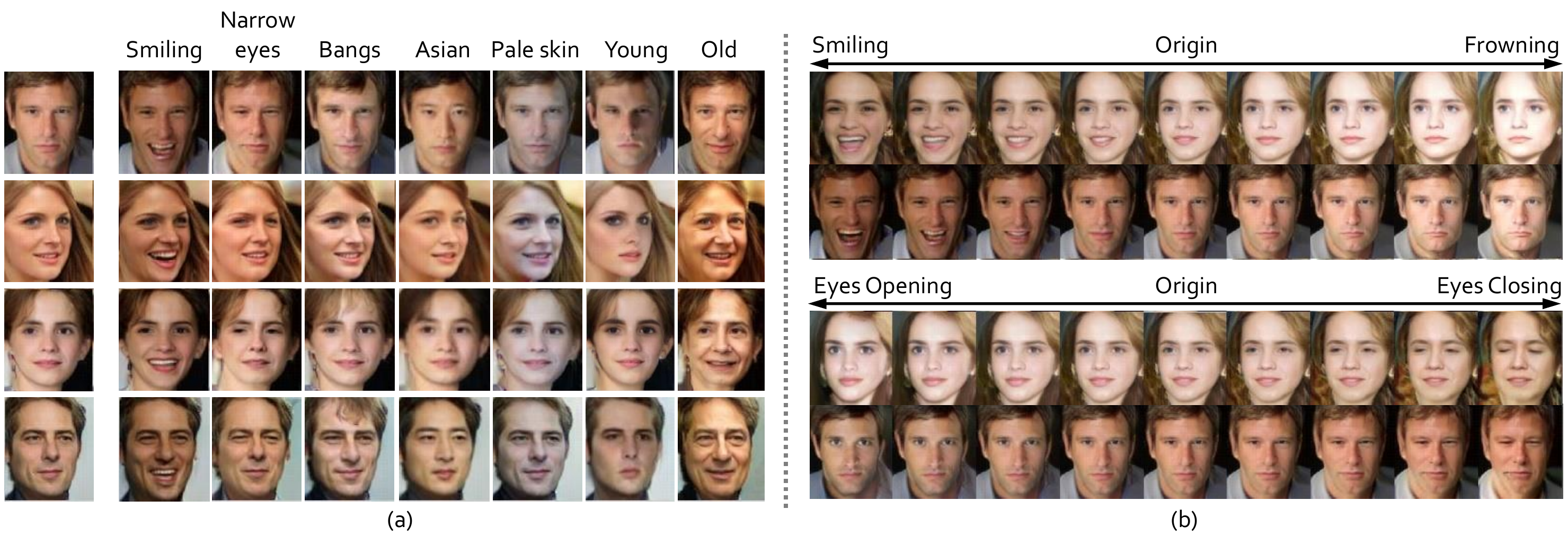}
\vspace{-0.1cm}
\caption{Results of (a) identity-aware attribute transfer and (b) identity-aware attribute interpolation. Zoom in for details.}
\label{fig:edit_several_attributes}
\vspace{-0.3cm}
\end{figure*}

\subsection{Face Identity Preserving}
\label{subsec:face_identity}

In this experiment we show the identity discriminability is preserved not only in the generated faces but also in the learned representations, with some interesting results. 
%
%

In addition to the MS-Celeb-1M dataset as mentioned in Sec~.\ref{sub:datasets_and_preprocessing}, we also compare their face verification results on a smaller CASIA-WebFace dataset~\cite{yi2014learning}.
It only contains $0.49$M images with around $10$K identities, approximately one-tenth of the scale of the MS-Celeb-1M dataset.
To further manifest the significance of the encoder-decoder structure compared to the encoder-only feature extraction scheme, we construct a baseline with the same encoder structure as included in the \modelname{} architecture, denoted as the \textit{Baseline} model.

As shown in Table~\ref{tb:compare_identity_ours}, \modelname{} achieves comparable performance with the Baseline model when trained on the MS-Celeb-1M dataset.
Furthermore, if trained on the WebFace dataset, it even outperforms the Baseline model.
This phenomenon occurs because the identity space may be biased towards some attributes due to the limited scale of the WebFace dataset.
For example, it is possible that the face images of a certain identity in the dataset always appear in the same pose or expression. In this case, such particular pose or expression is likely to be used to define this identity, and hence it will be falsely encoded in the identity-related feature.
Because the \modelname{} model disentangles a face representation into an identity-distilled feature and a complementary identity-dispelled feature, it owns a superiority over the baseline model in the task of identity verification.
In contrast, when trained on the MS-Celeb-1M dataset which has a sufficiently large scale, the baseline model is able to correctly extract identity-related information.
Comparison of results between the \modelname{} model and the prior arts is plotted in Fig.~\ref{fig:fig_1_intro} (c).

\begin{table}[t]
\center
\footnotesize{
\begin{tabular}{M{1.6cm}M{0.9cm}M{0.9cm}M{0.9cm}M{0.9cm}}
\hline
\multirow{2}{*}{\textbf{Method}} & \multicolumn{2}{c}{\textbf{MS-Celeb-1M}} & \multicolumn{2}{c}{\textbf{WebFace}} \\
\cline{2-5}
 & \textbf{Acc} & \textbf{TPR} & \textbf{Acc} & \textbf{TPR}\\
\hline
Baseline & 99.816 & 99.73 & 98.93 & 94.87 \\
\modelname{} & 99.80 & 99.40 & 99.25 & 96.80 \\
\hline
\end{tabular}
}
\vspace{+1mm}
\caption{Comparison of results on face identity verification.}
\label{tb:compare_identity_ours}
\vspace{-2mm}
\end{table}

\vspace{-0.1cm}
\subsection{Face Attribute Recognition}
\label{subsec:face_attribute}
\vspace{-0.2cm}

\begin{table}[t]
\center
\footnotesize{
\begin{tabular}{M{1.6cm}|M{0.9cm}M{0.9cm}|M{0.9cm}}
\hline
\textbf{Dataset} & \cite{liu2015} & \cite{zhang2014panda} & \modelname{} \\
\hline
LFWA & 83.85 & 81.03 & \underline{83.16} \\
CelebA & 87.30 & 85.43 & \underline{87.82} \\
\hline
\end{tabular}
}
\vspace{0.1cm}
\caption{Comparison of results on face attribute recognition.}
\label{tb:compare_attribute}
\vspace{-0.4cm}
\end{table}

We compare the proposed framework and two methods \cite{liu2015, zhang2014panda} with supervision on face attribute.
Performance is evaluated on two commonly employed datasets, \ie,~LFWA~\cite{liu2015} and CelebA~\cite{liu2015}.
As shown in Table~\ref{tb:compare_attribute}, although the proposed method is entirely unsupervised in terms of face attribute, it achieves results comparable with the supervised methods.

In Fig.~\ref{fig:fig_1_intro} (b),
 we visualize the 2D embedding of all faces with attributes in LFWA dataset, based on the Barnes-Hut \textit{t}-SNE method \cite{Maaten2017Visualizing}. 
We can observe that with features extracted from \modelname{}, the 2D embedding space can be automatically partitioned by either attributes or identities.
Note that the attributes do not follow category boundaries.
There are overlapping classification boundaries for different identity-aware attributes such as \textit{sex} and \textit{race}, while the face images for one identity are densely clustered.
%

\vspace{-0.1cm}
\subsection{Face Editing}
\label{subsec:face_editing}
\vspace{-0.2cm}

We show that the proposed method presents superior performances on semantic face editing.\footnote{For more examples please refer to Fig.~\ref{fig:moresample} in appendix.}
We take several face images from the LFW dataset and reconstruct them with modification on different attributes as well as identities.

\vspace{+1mm}
\noindent\textbf{Identity-aware Attribute Editing.}
Fig.~\ref{fig:edit_several_attributes} (a) shows several portraits with one attribute changed at a time.
Our model alters the attributes with well-preserved naturalness and identity.
For either local (\eg,~\textit{smiling}, \textit{narrow eyes}, and \textit{bangs}) or global attributes (\eg,~\textit{Asian} and \textit{age}), the model has successfully disentangled identity-distilled and identity-dispelled features.
For example, even if the transformation of race certainly disturbs identity, almost all the identity-irrelevant attributes such as hair style, facial expression, and background color are well preserved.
In the last column, the proposed model even synthesizes the unseen teeth and tongues when editing a portrait to \textit{smile} and generates wrinkles when altering a person's age towards \textit{older}.

\vspace{+1mm}
\noindent\textbf{Identity-aware Attribute Interpolation.}
Interpolation is performed by changing the weight of an attribute, which renders face images with different magnitudes of that attribute as shown in Fig.~\ref{fig:edit_several_attributes} (b).
Our model enables smooth and natural change of either a female face from \textit{smiling} to \textit{frowning} or eyes of a male from being open to being closed.

\vspace{+1mm}
\noindent\textbf{Identity Transfer.}
To further illustrate the compactness of the convex feature space, a face image is gradually changed from one identity to another in the last row of Fig.~\ref{fig:fig_1_intro} (a).
It shows a smooth transition from a smiling female to a frowning male with the hair style gradually changed as well.
More results are shown in Fig.~\ref{fig:clinton2trump}.

\begin{figure}
\centering
\includegraphics[width=\linewidth]{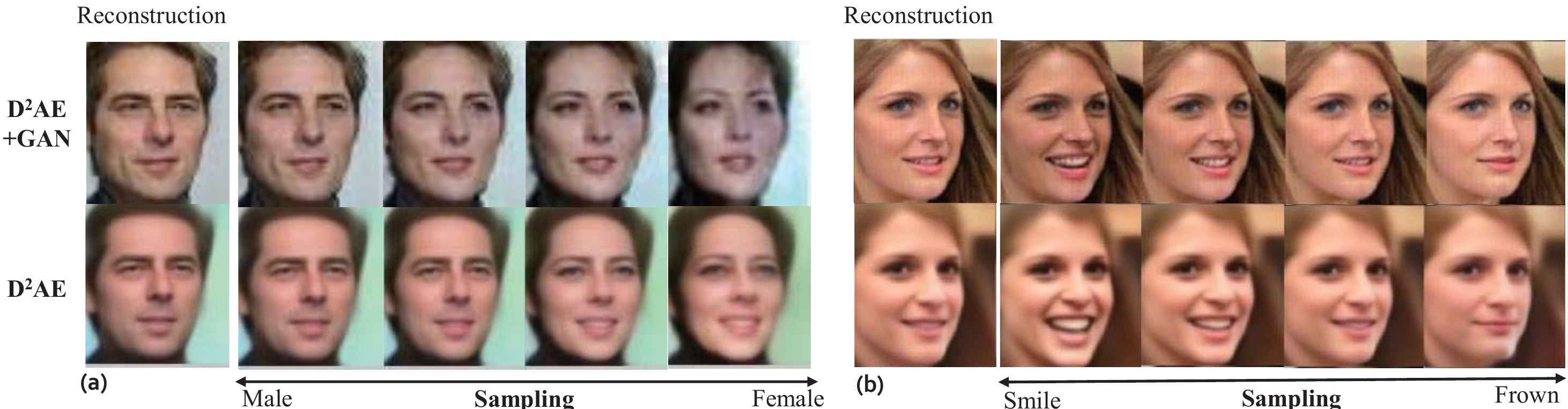}
\caption{Identity-aware attribute transfer using the proposed model and its extension based on the GAN model.}
\label{fig:edit_vae_vs_gan}
\vspace{-0.2cm}
\end{figure}

\vspace{+1mm}
\noindent\textbf{Extension toward GANs.}
We find that the proposed model can be safely incorporated with the generative adversarial networks (GANs)~\cite{goodfellow2014generative}, by switching the reconstruction loss $\mathcal{L}_\mathcal{X}$ to the adversarial loss from an additional discriminator.
The reconstructed face images contain more realistic details and noises as shown in Fig.~\ref{fig:edit_vae_vs_gan}.

\vspace{-0.2cm}
\section{Conclusion}
\label{sec:conclusion}
\vspace{-0.2cm}

The proposed \modelname{} disentangles the face representation into two orthogonal streams with novel adversarial supervision. Features in the two streams completely represent the information in the whole face, which are highly distinctive and densely distributed in a convex latent space. The learned features are ready for various applications such as face verification, attribute prediction and face editing, where the model all achieves state-of-the-art performances.
%
%

\newcommand{\subfigwidth}{0.23}
\newcommand{\subfigheight}{0.245}
\begin{figure*}
\centering
\bgroup
\setlength\tabcolsep{1.5pt} 
\begin{tabular}{cccc}
    \includegraphics[width=\subfigwidth\linewidth,height=\subfigheight\textheight]{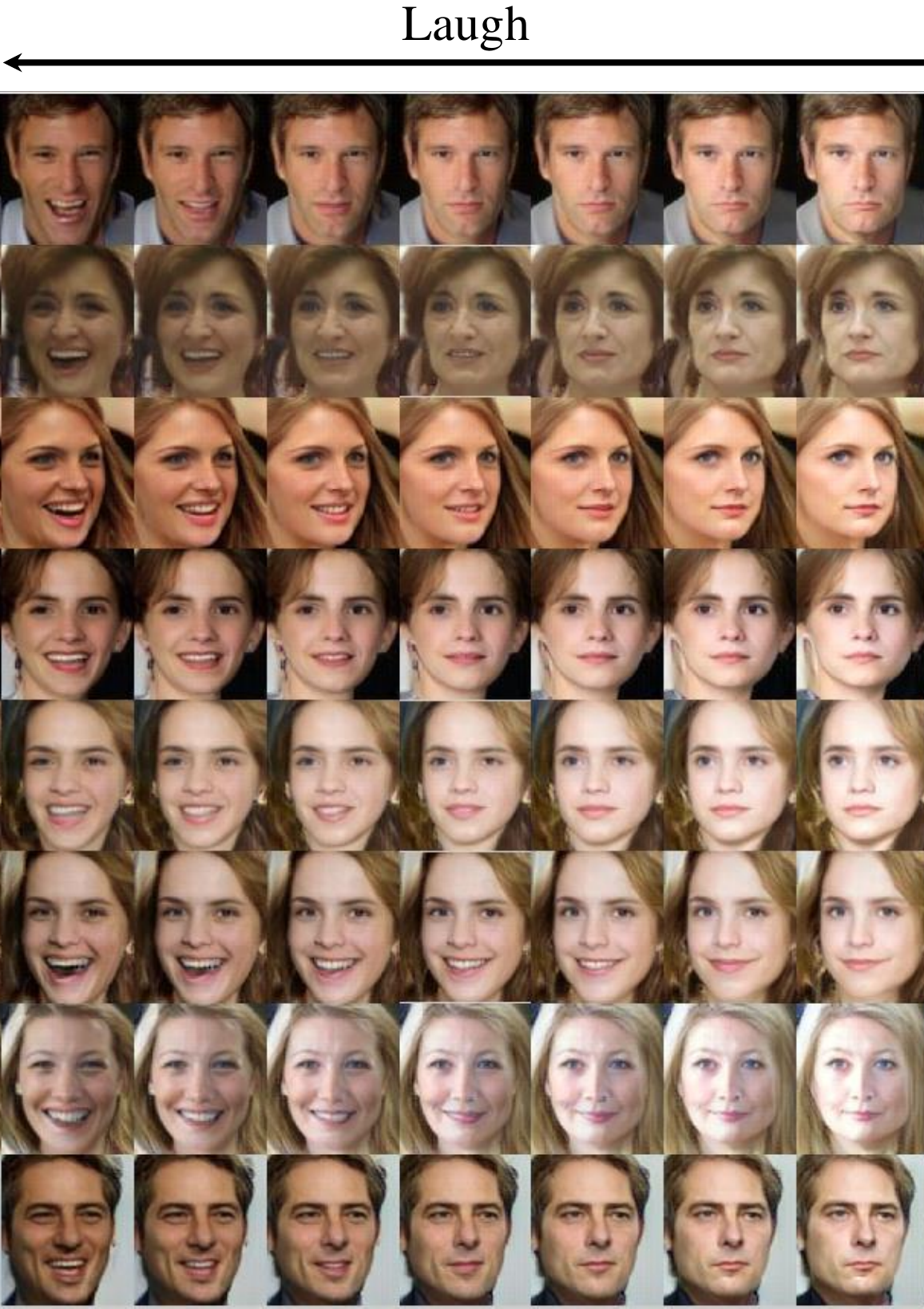} &
    \includegraphics[width=\subfigwidth\linewidth,height=\subfigheight\textheight]{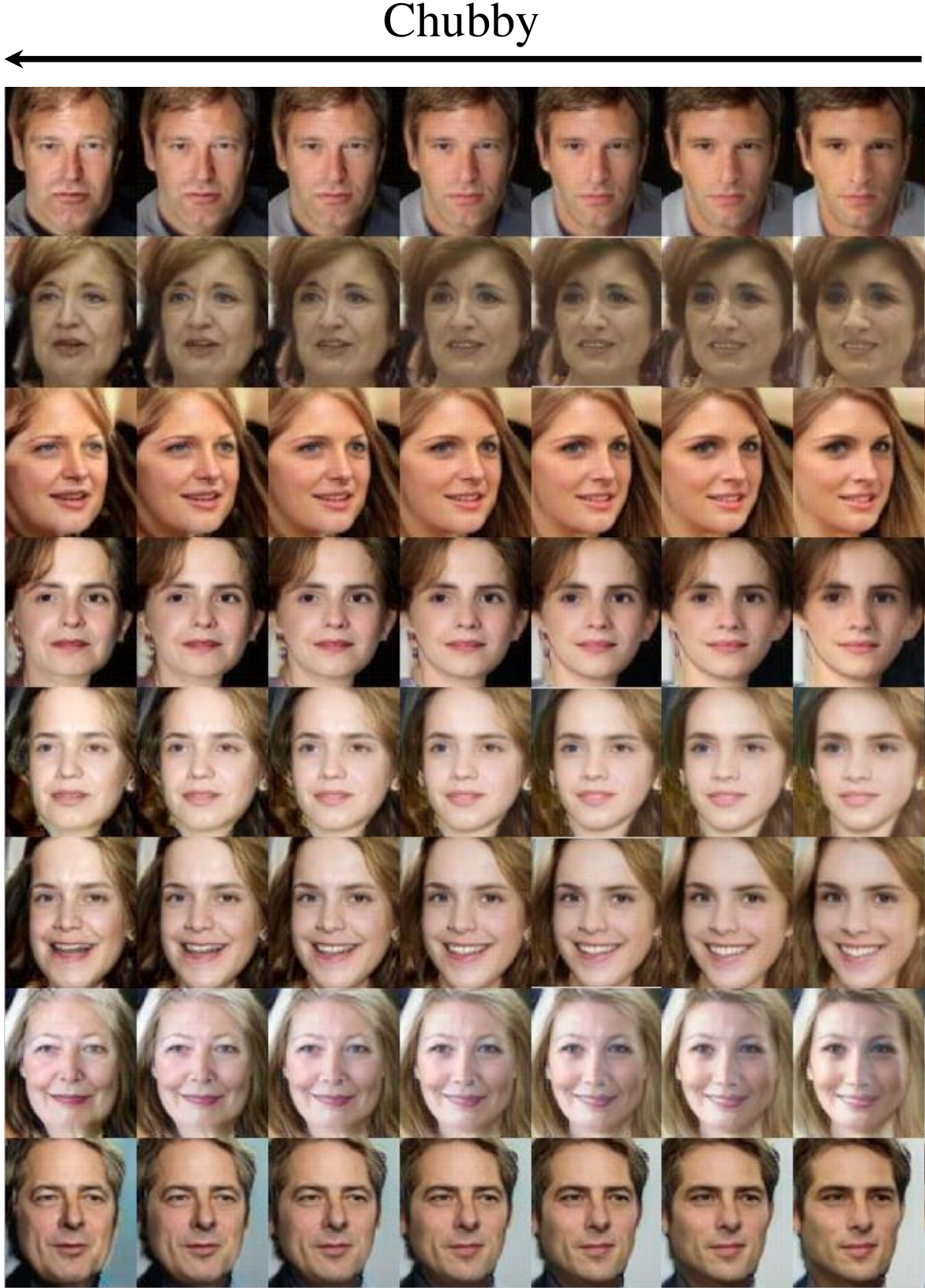} &
    \includegraphics[width=\subfigwidth\linewidth,height=\subfigheight\textheight]{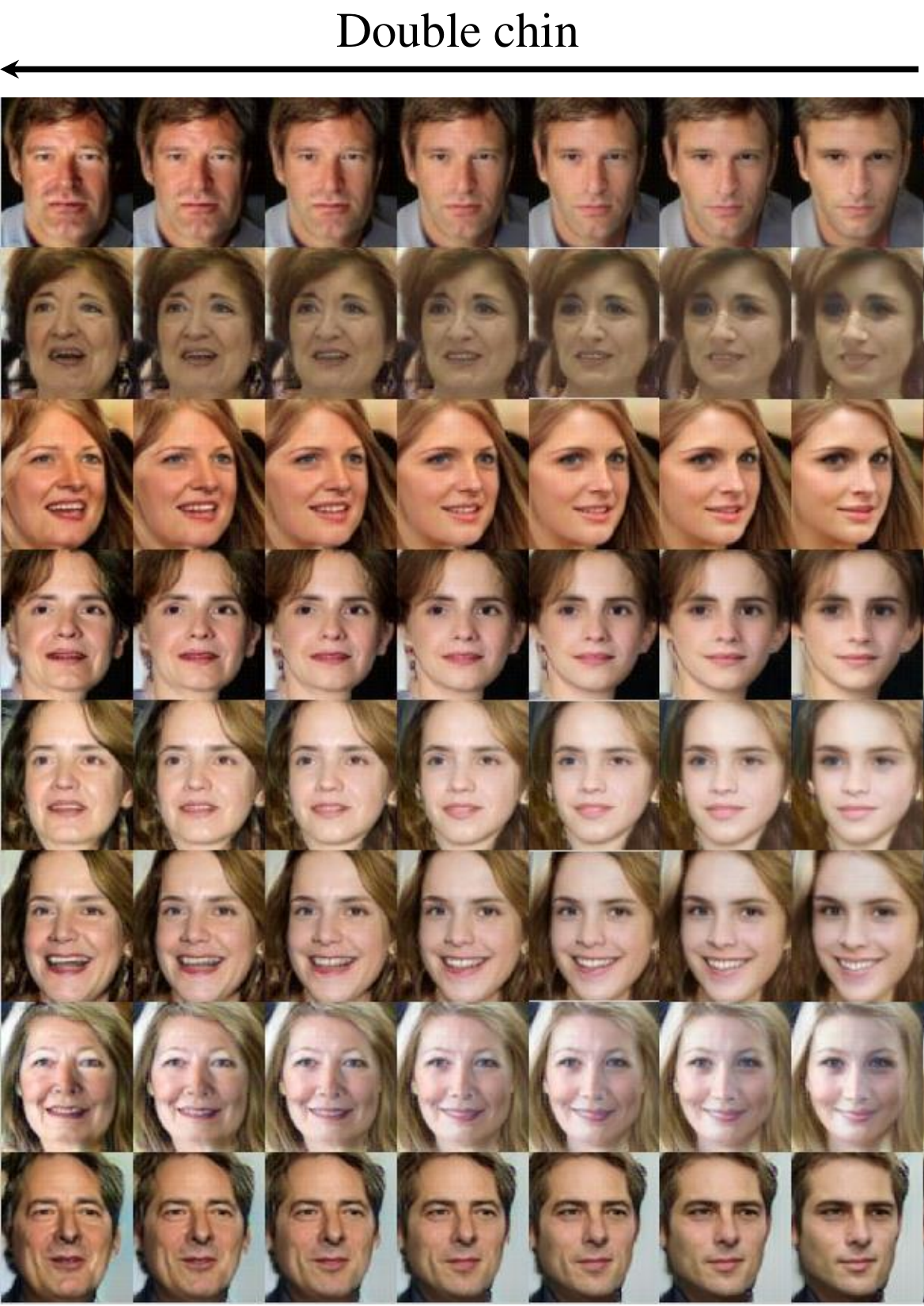} &
    \includegraphics[width=\subfigwidth\linewidth,height=\subfigheight\textheight]{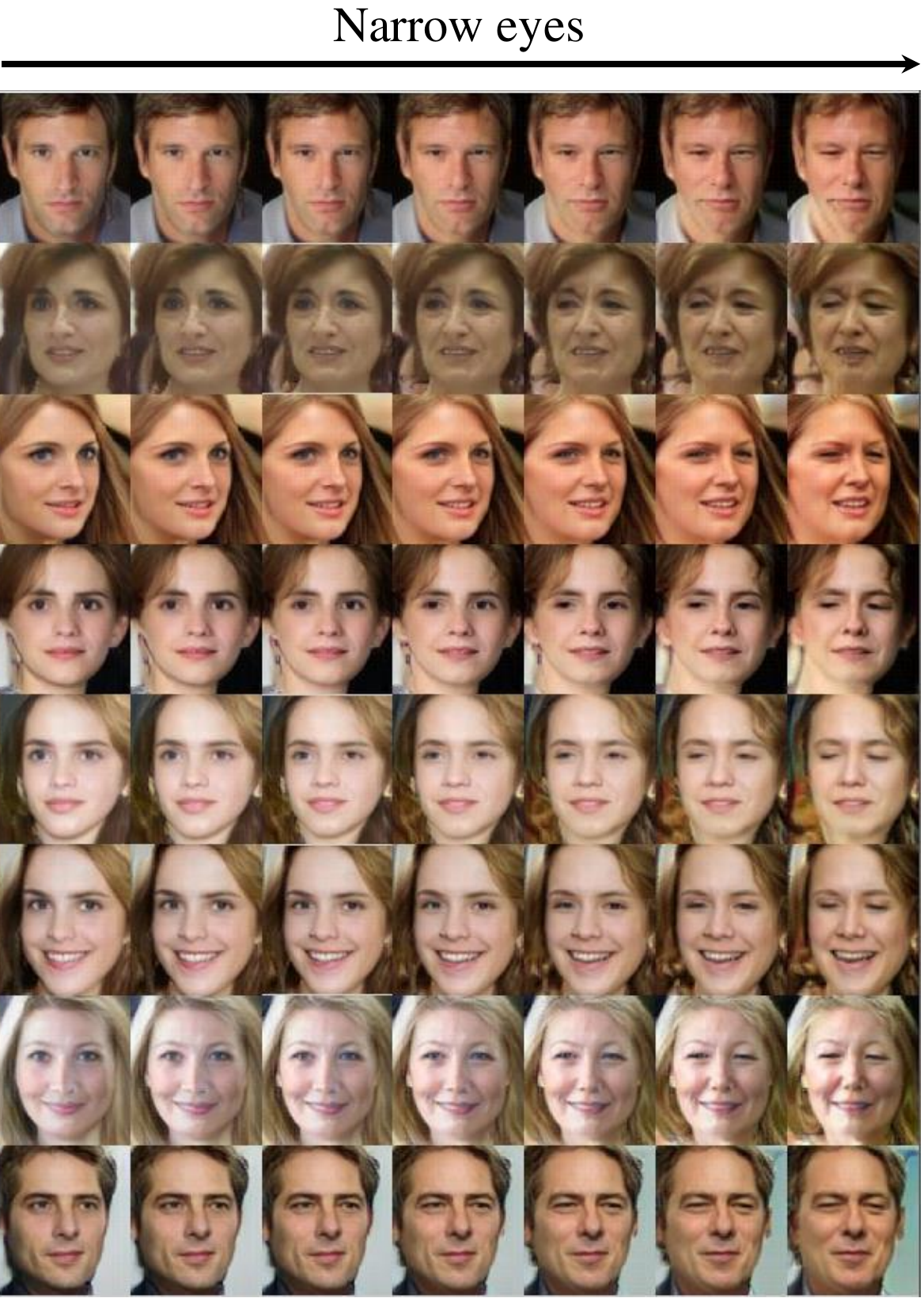} \\	
   \includegraphics[width=\subfigwidth\linewidth,height=\subfigheight\textheight]{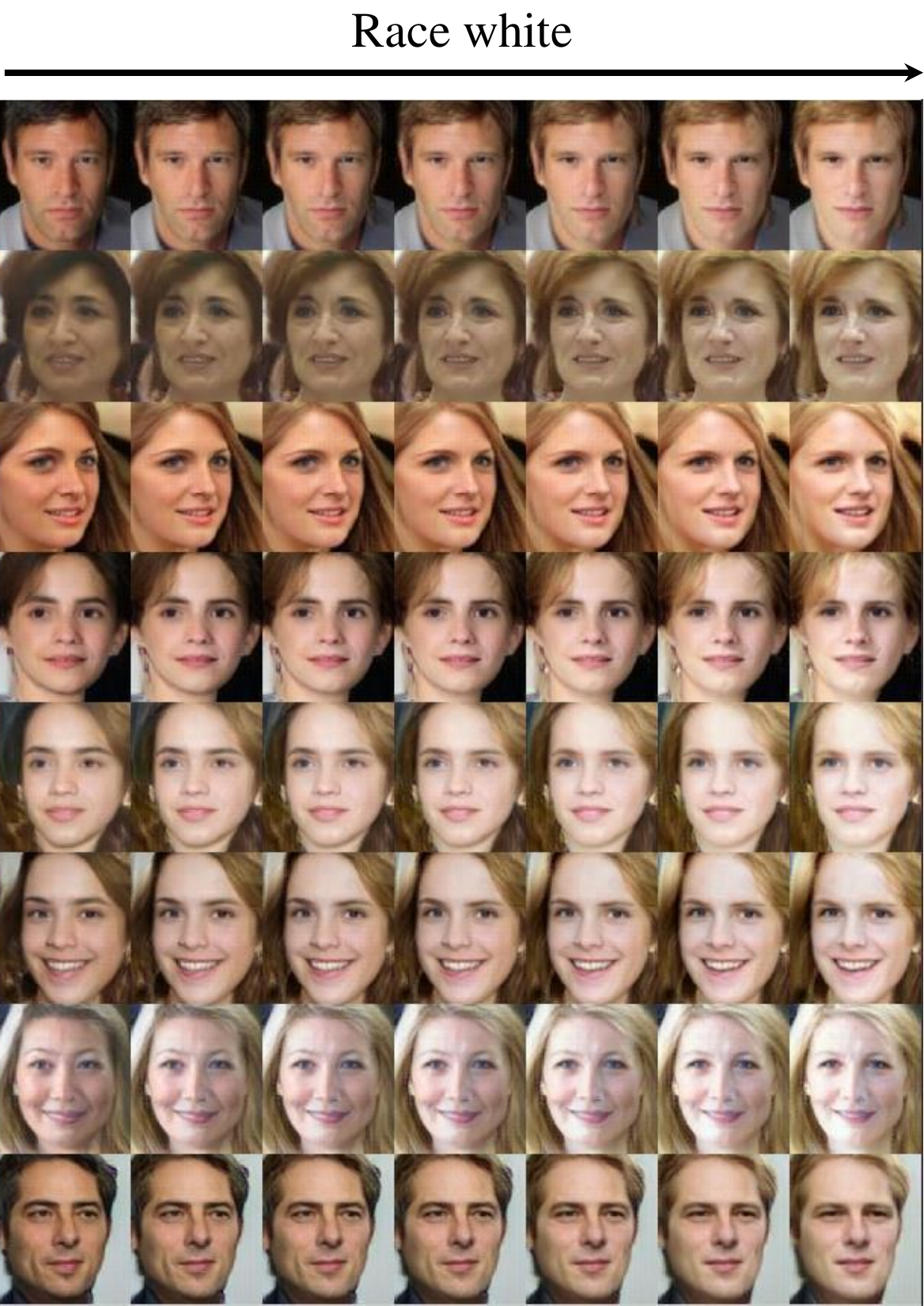} &
    \includegraphics[width=\subfigwidth\linewidth,height=\subfigheight\textheight]{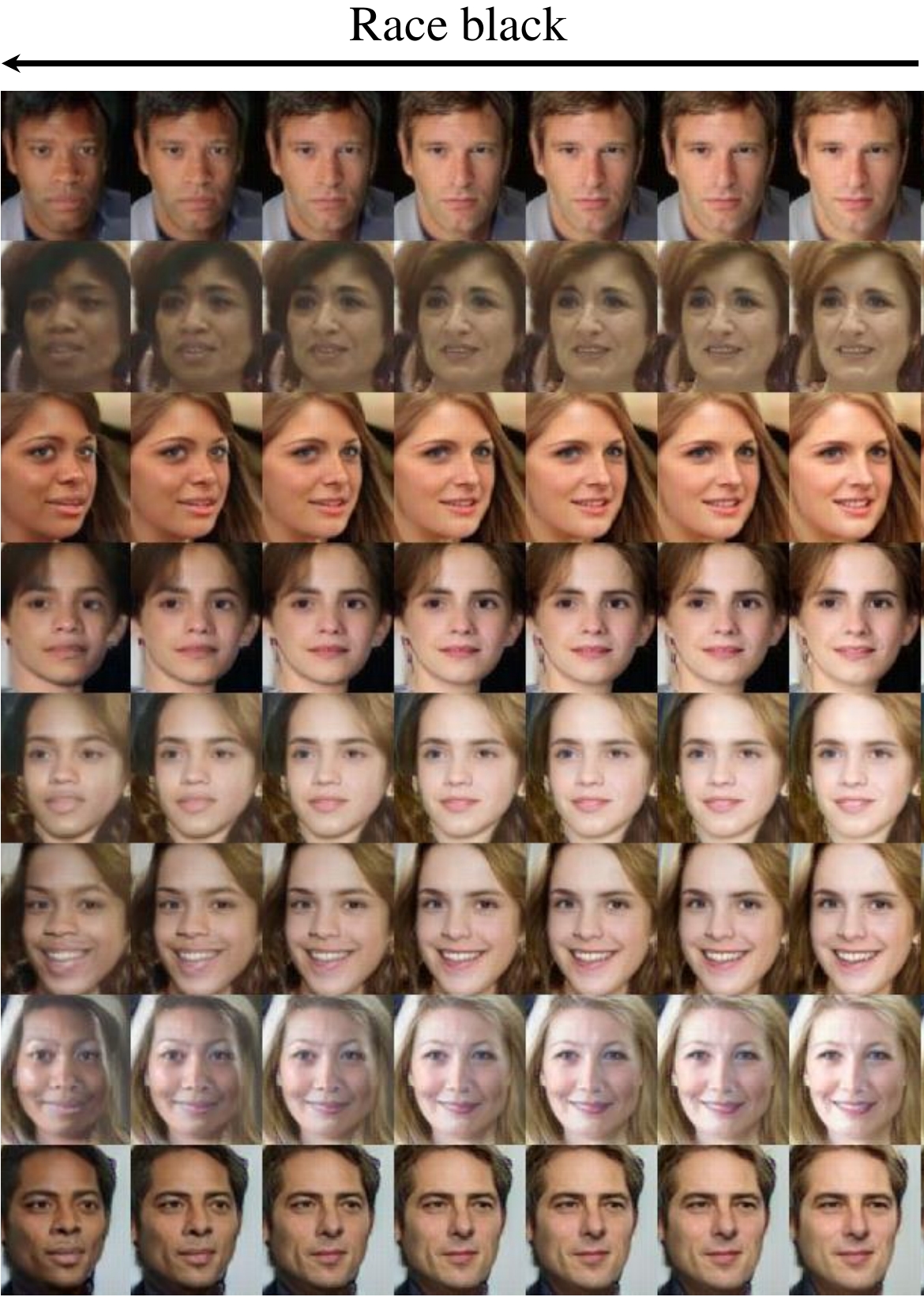} &
    \includegraphics[width=\subfigwidth\linewidth,height=\subfigheight\textheight]{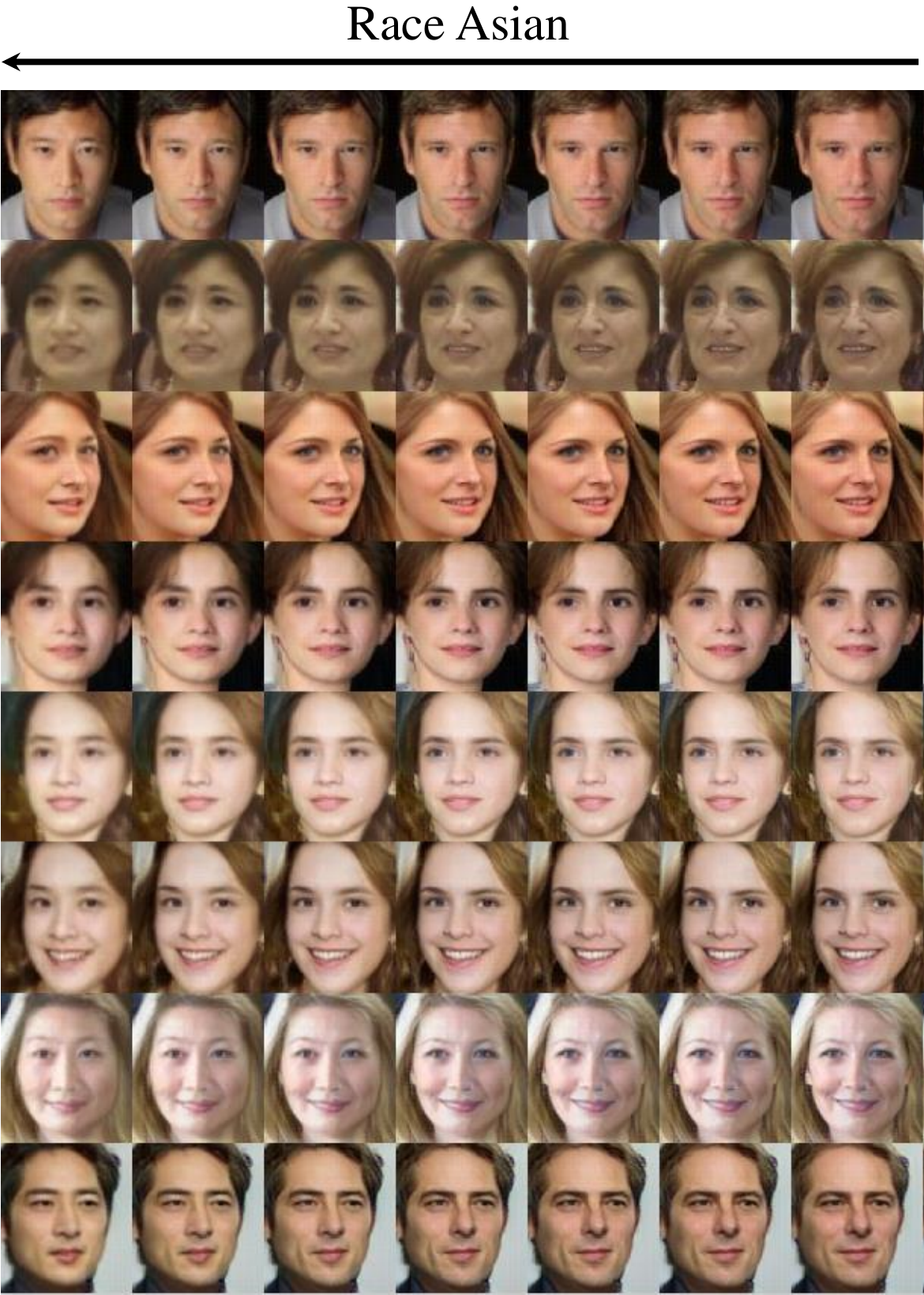} &
    \includegraphics[width=\subfigwidth\linewidth,height=\subfigheight\textheight]{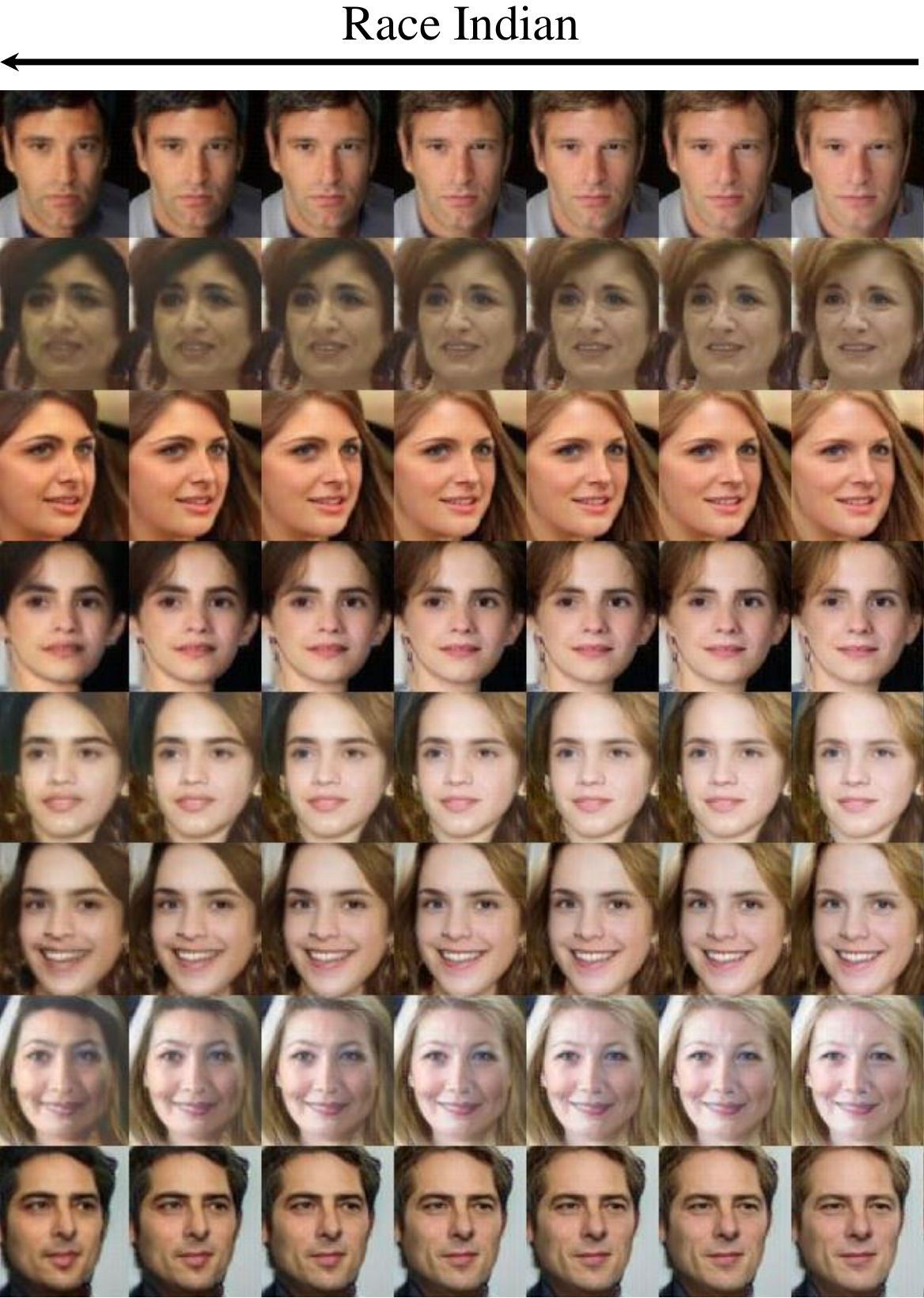}  \\
    \includegraphics[width=\subfigwidth\linewidth,height=\subfigheight\textheight]{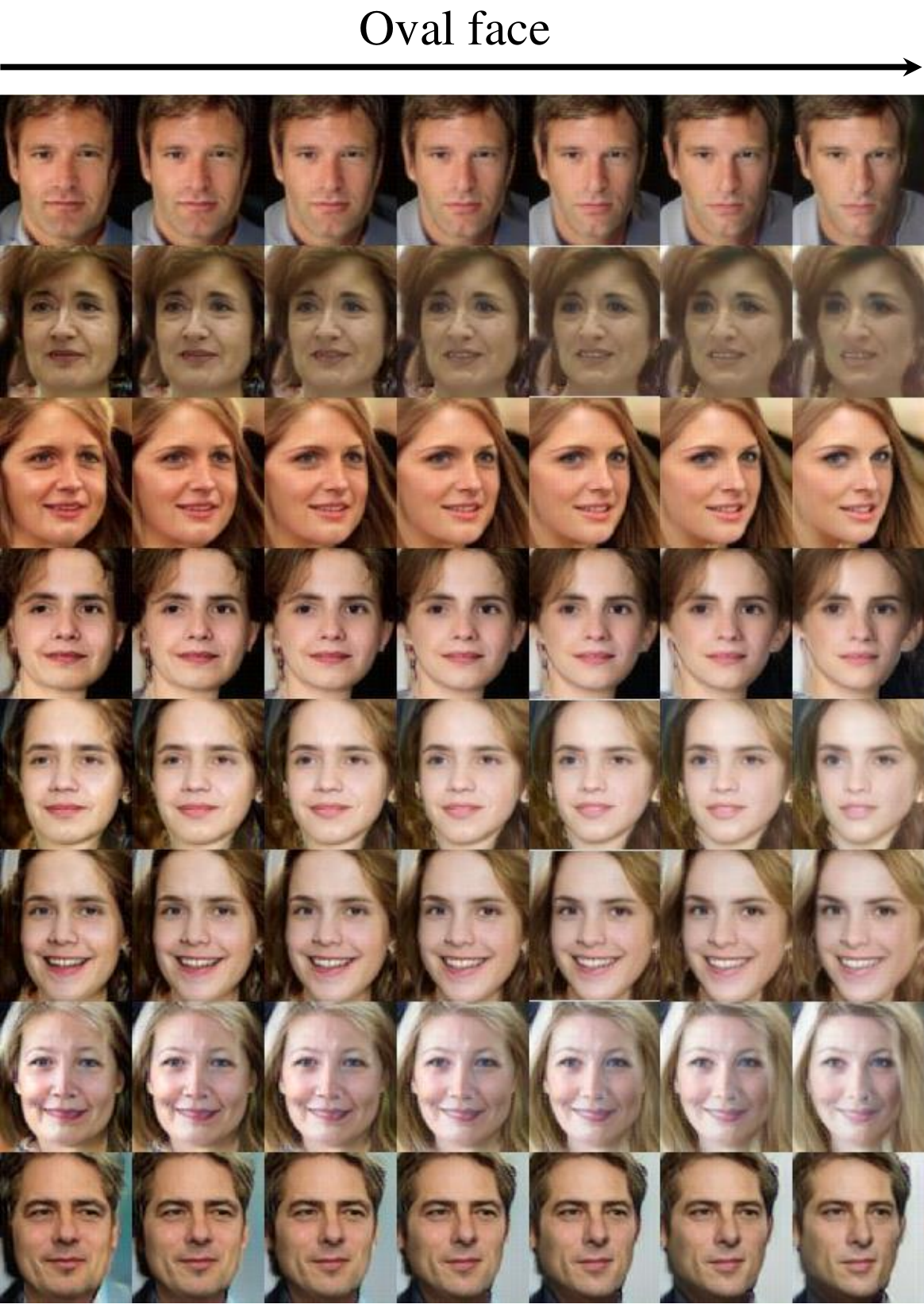} &
    \includegraphics[width=\subfigwidth\linewidth,height=\subfigheight\textheight]{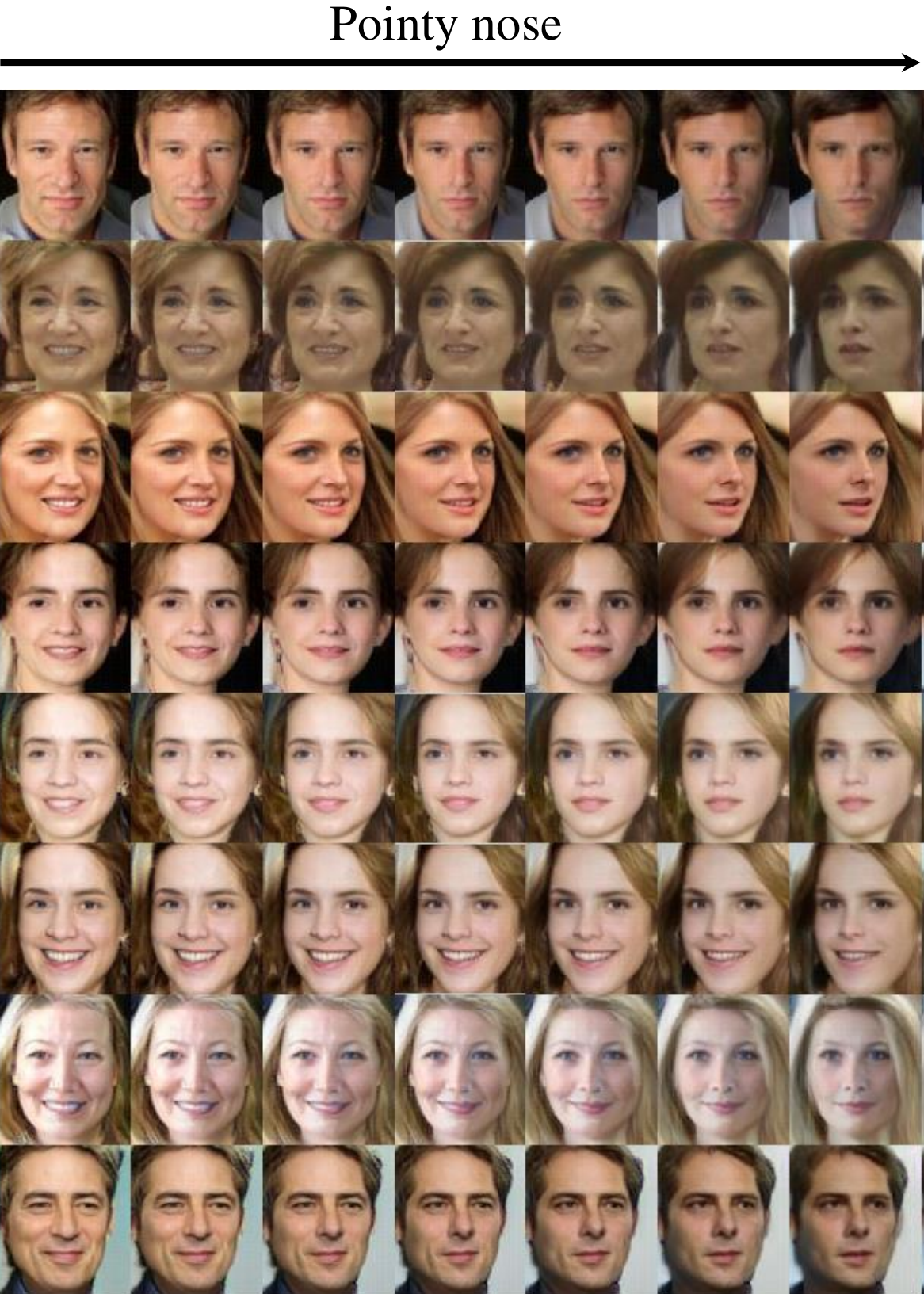} &
    \includegraphics[width=\subfigwidth\linewidth,height=\subfigheight\textheight]{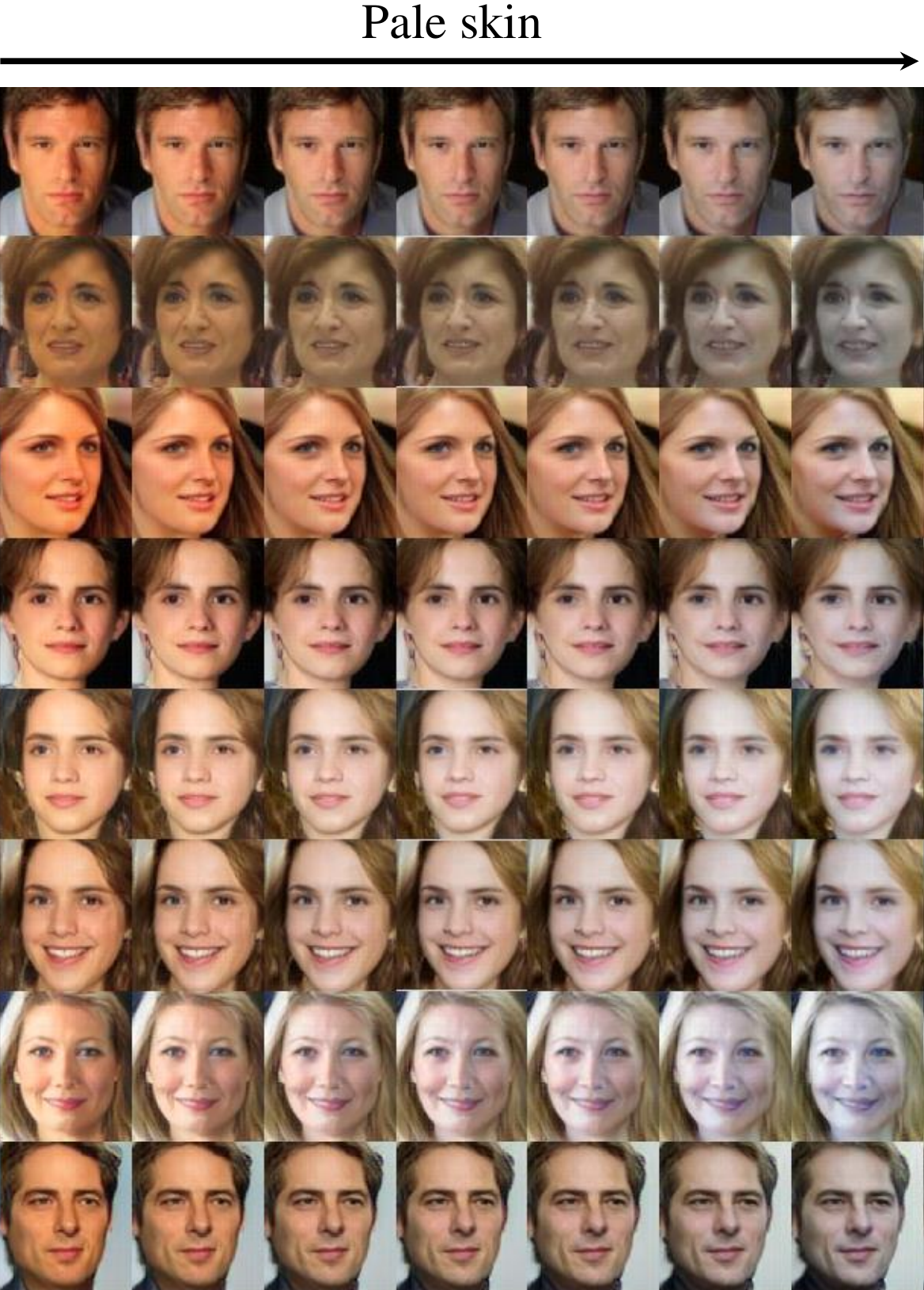} &
    \includegraphics[width=\subfigwidth\linewidth,height=\subfigheight\textheight]{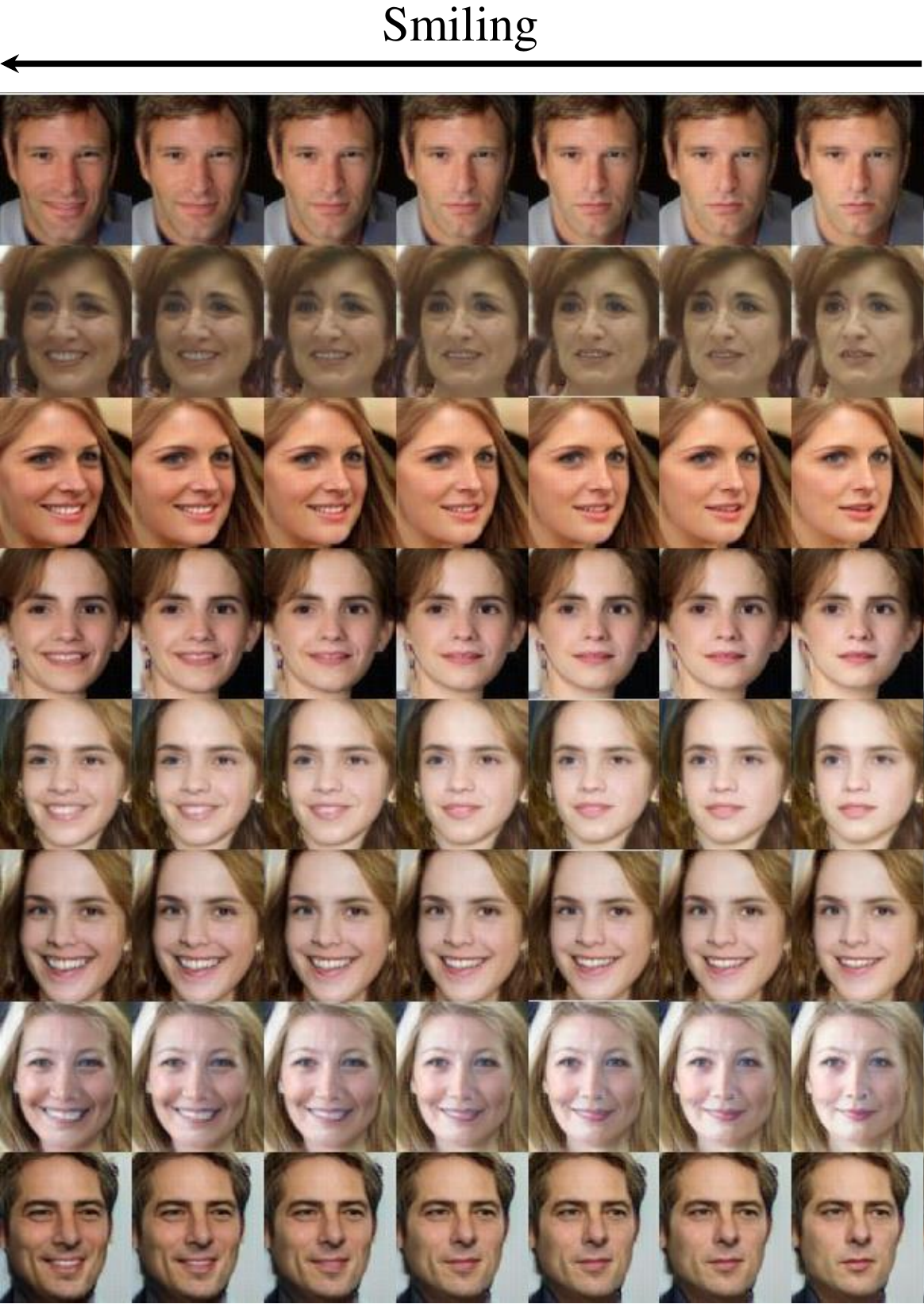} \\
    \includegraphics[width=\subfigwidth\linewidth,height=\subfigheight\textheight]{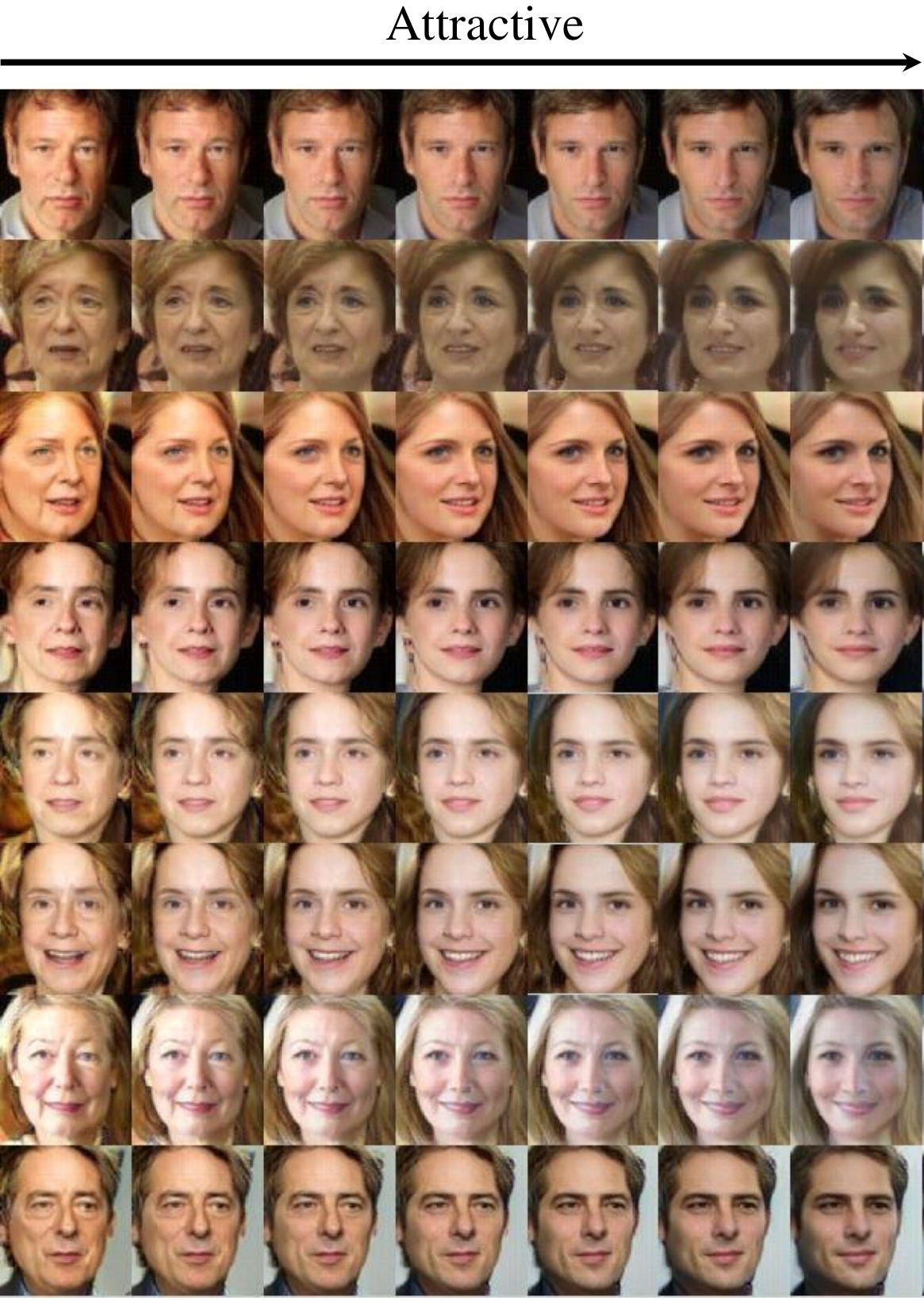} &
   \includegraphics[width=\subfigwidth\linewidth,height=\subfigheight\textheight]{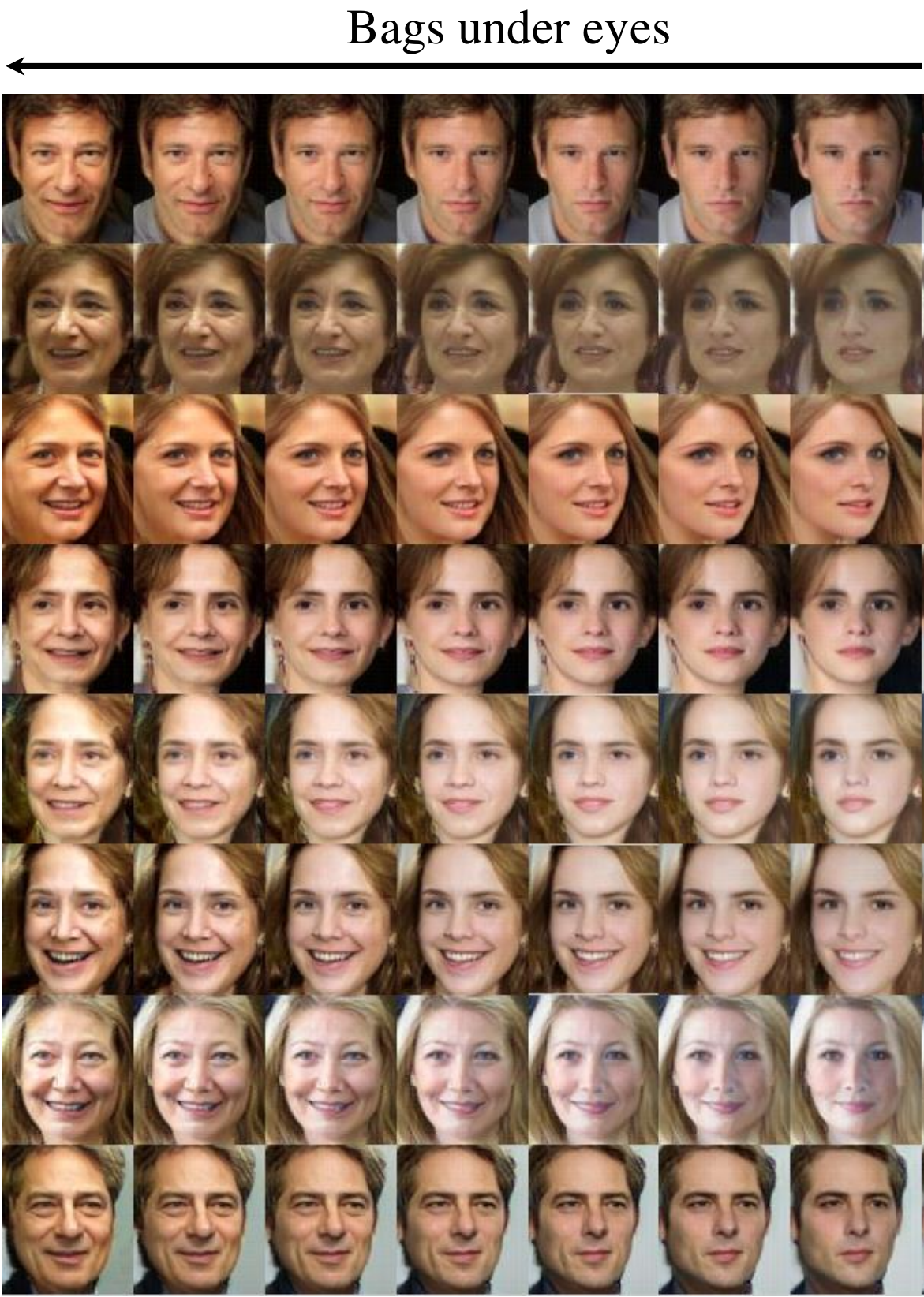} &
   \includegraphics[width=\subfigwidth\linewidth,height=\subfigheight\textheight]{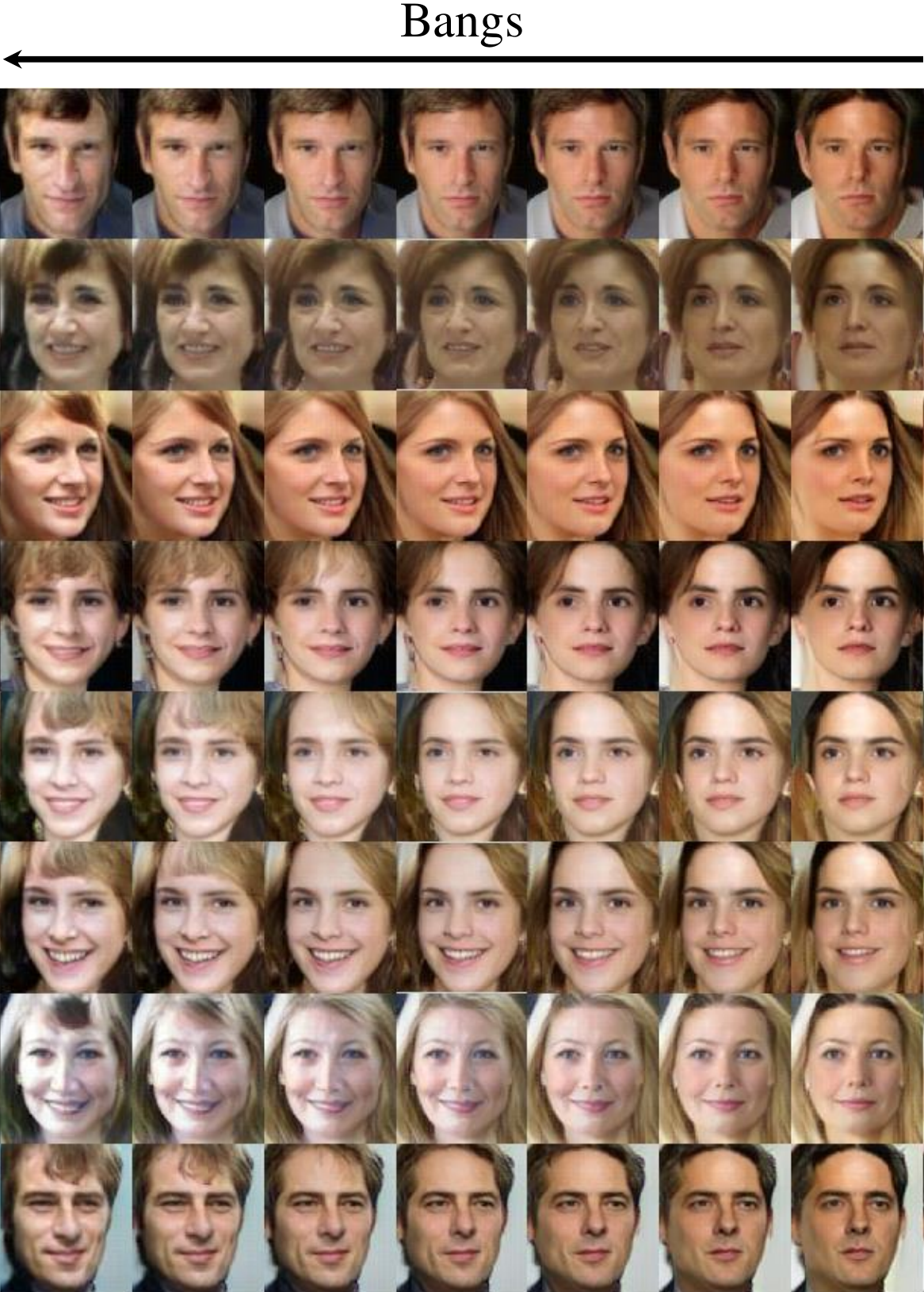} &
   \includegraphics[width=\subfigwidth\linewidth,height=\subfigheight\textheight]{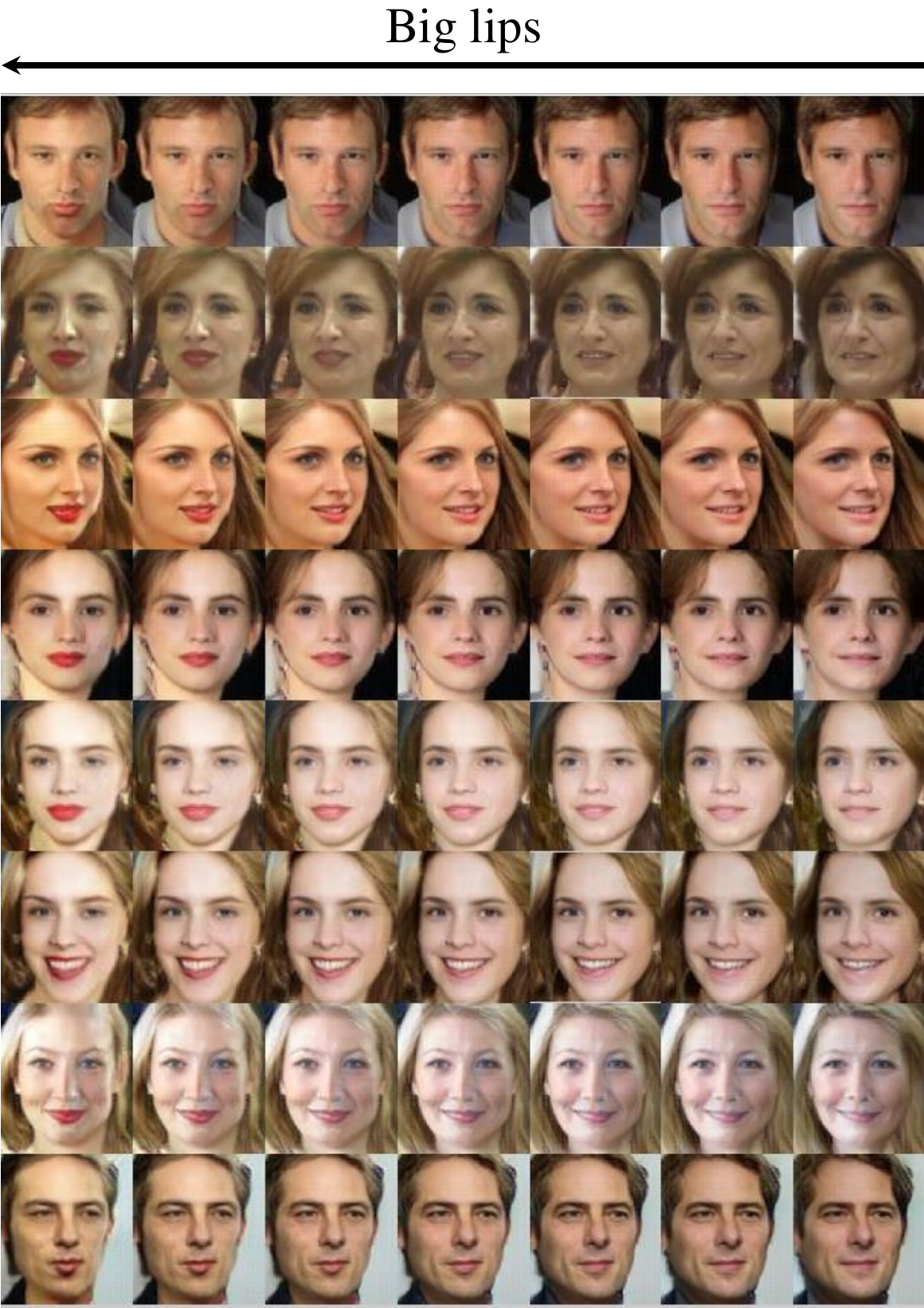}
\end{tabular}
\egroup
\caption{More results for Identity-aware attribute aditing.}
\label{fig:moresample}
\end{figure*}

\appendix  

\makeatletter
\def\hlinew#1{%
  \noalign{\ifnum0=`}\fi\hrule \@height #1 \futurelet
   \reserved@a\@xhline}
\makeatother

\section{Training Details}
\label{sec:training_details}
\subsection{Network Structure}
\textbf{Encoder Module:} We use Inception-ResNet~\cite{szegedy2017inception} as the backbone of $E_{\boldsymbol\theta_\text{enc}}$. The input size is modified to $235\times235$ so that the size of the feature maps before the final \texttt{AvePool} layer is $6\times 6$. We further replace the \texttt{AvePool} layer with two 256-dim \texttt{FC} layers, \ie, $B_{\boldsymbol\theta_\mathcal{T}}$ and $B_{\boldsymbol\theta_\mathcal{P}}$. During training process, we first pre-train this model in a simple classification manner on the corresponding face datasets. Then the pre-trained model is used as an initial model for training \modelname.

\textbf{Decoder Module:} As shown in Table~\ref{tab:decoder}, there are 20 \texttt{conv} layers, 6 \texttt{upsample} layers and one \texttt{deconv} layer in the decoder. Parameters of \texttt{conv} and \texttt{deconv} layers are initialized with \texttt{Gaussian}. Note that the decoder produces images with size of $256\times 256$ and is supervised by the reconstruction loss with the scaled input image.

\begin{table}[h]
\centering
\caption{Structure details of decoder.}
\label{tab:decoder}
\begin{tabular}{c|c|c|c}
\hlinew{1.5pt}
Layer Type & Dim. Out                & Kernal & Num. \\ \hline \hline
Concat     & $1\times1\times512$     & -           & -                \\ \hline
Deconv     & $4\times4\times512$     & 5           & 1                \\ \hline
Conv       & $4\times4\times512$     & 3           & 2                \\ \hline
Upsample   & $8\times8\times512$     & -           & 2                \\ \hline
Conv       & $8\times8\times512$     & 3           & 3                \\ \hline
Upsample   & $16\times16\times512$   & -           & 1                \\ \hline
Conv       & $16\times16\times256$   & 3           & 4                \\ \hline
Upsample   & $32\times32\times256$   & -           & 1                \\ \hline
Conv       & $32\times32\times256$   & 3           & 3                \\ \hline
Upsample   & $64\times64\times256$   & -           & 1                \\ \hline
Conv       & $64\times64\times128$   & 3           & 3                \\ \hline
Upsample   & $128\times128\times128$ & -           & 1                \\ \hline
Conv       & $128\times128\times64$  & 3           & 2                \\ \hline
Upsample   & $256\times256\times64$  & -           & 1                \\ \hline
Conv       & $256\times256\times32$  & 3           & 1                \\ \hline
Conv       & $256\times256\times3$   & 1           & 1                \\ \hlinew{1.5pt}
\end{tabular}
\end{table}

\begin{figure}[h]
\centering
	\includegraphics[width=.95\linewidth]{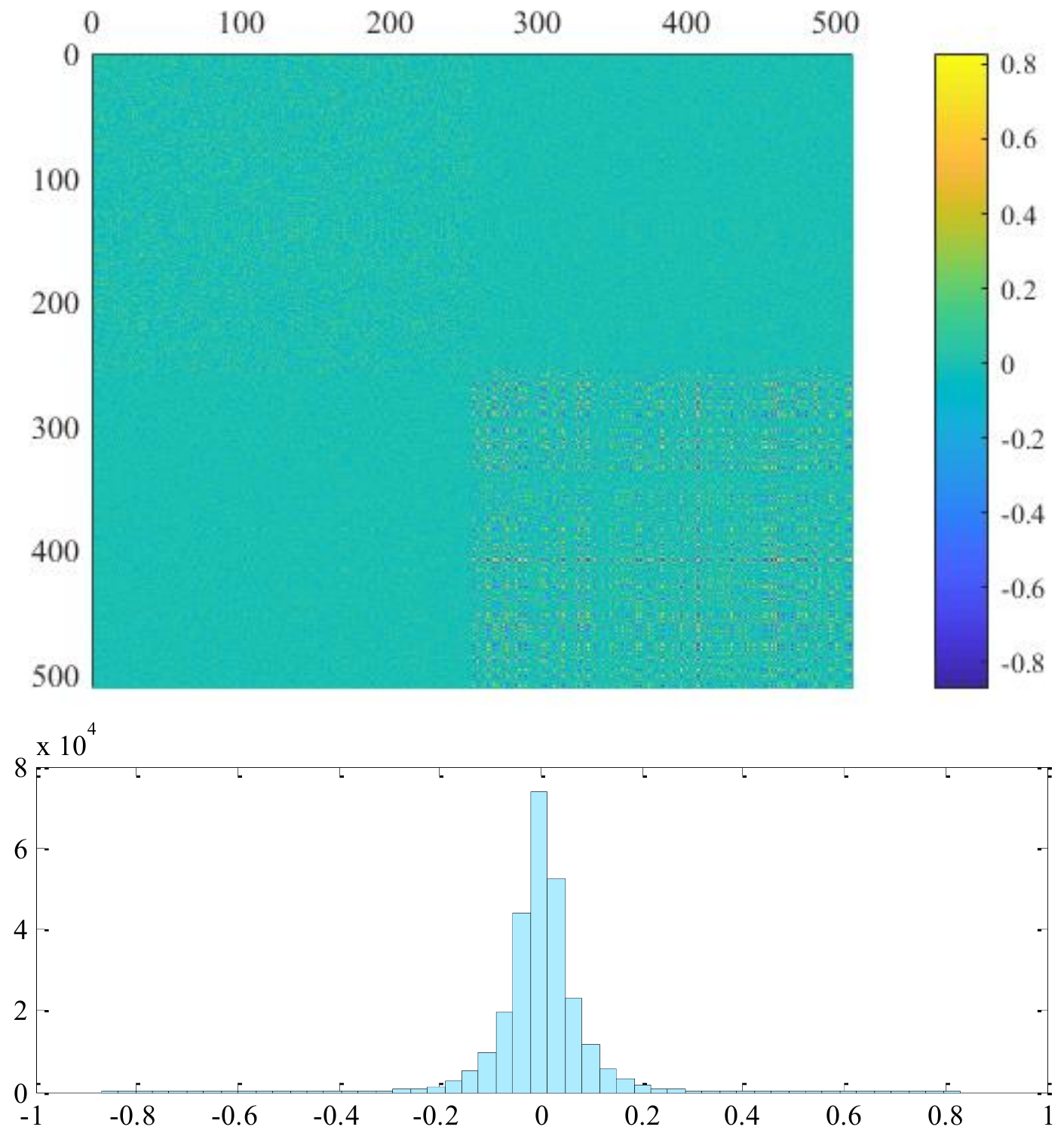}
	\caption{\textbf{Top}: correlation between 512 channels of features from \modelname-$\mathcal{P}$ and \modelname-$\mathcal{T}$ (diagonal elements are set to zero). \textbf{Bottom}: distribution of correlation coefficients.}
  	\label{fig:correlation}
\end{figure}

\section{Interpretation for Learned Gaussian Space}
\label{sec:interpretation}

\begin{figure}
\centering
	\includegraphics[width=.95\linewidth]{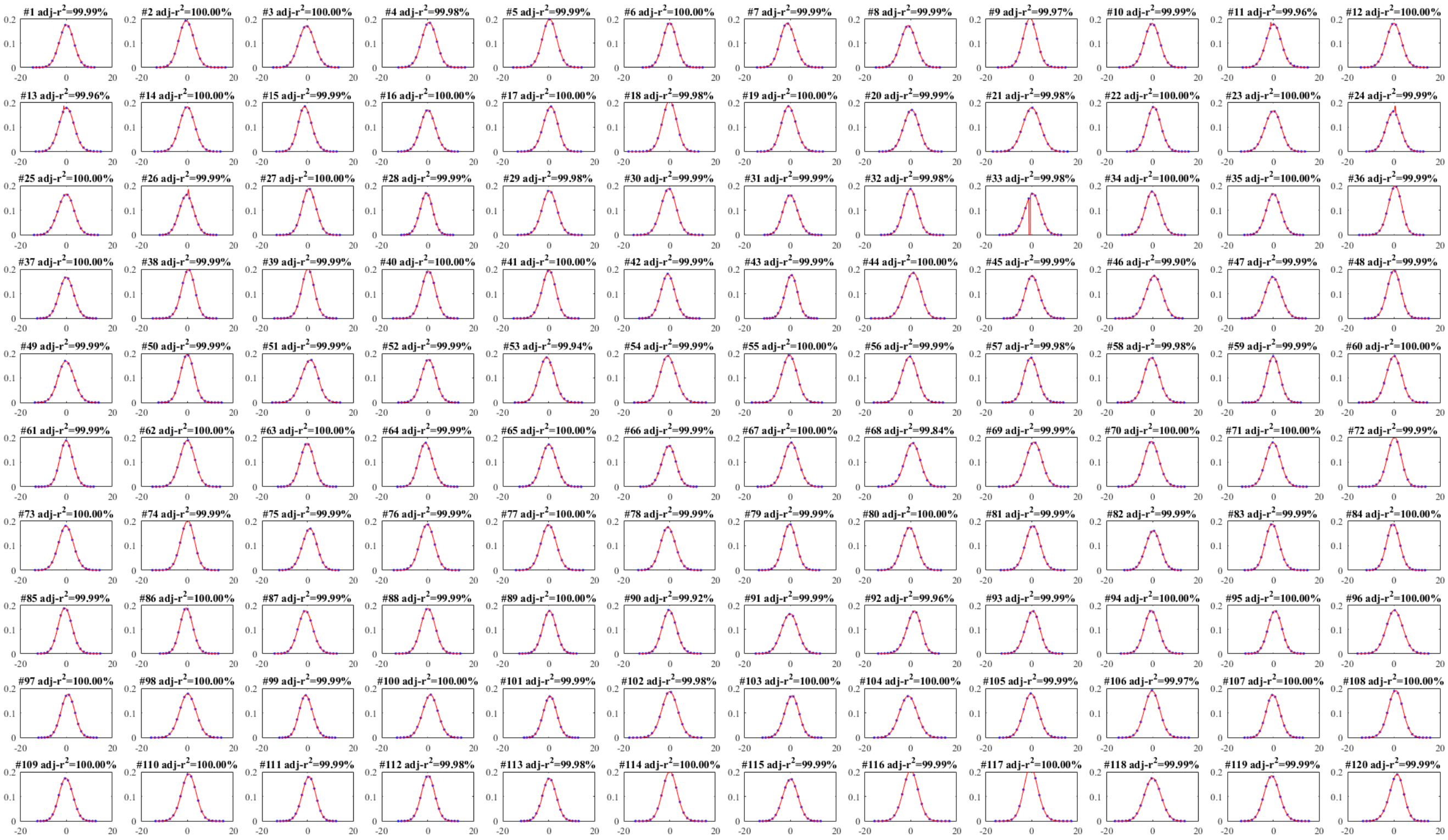}
  	\includegraphics[width=.95\linewidth]{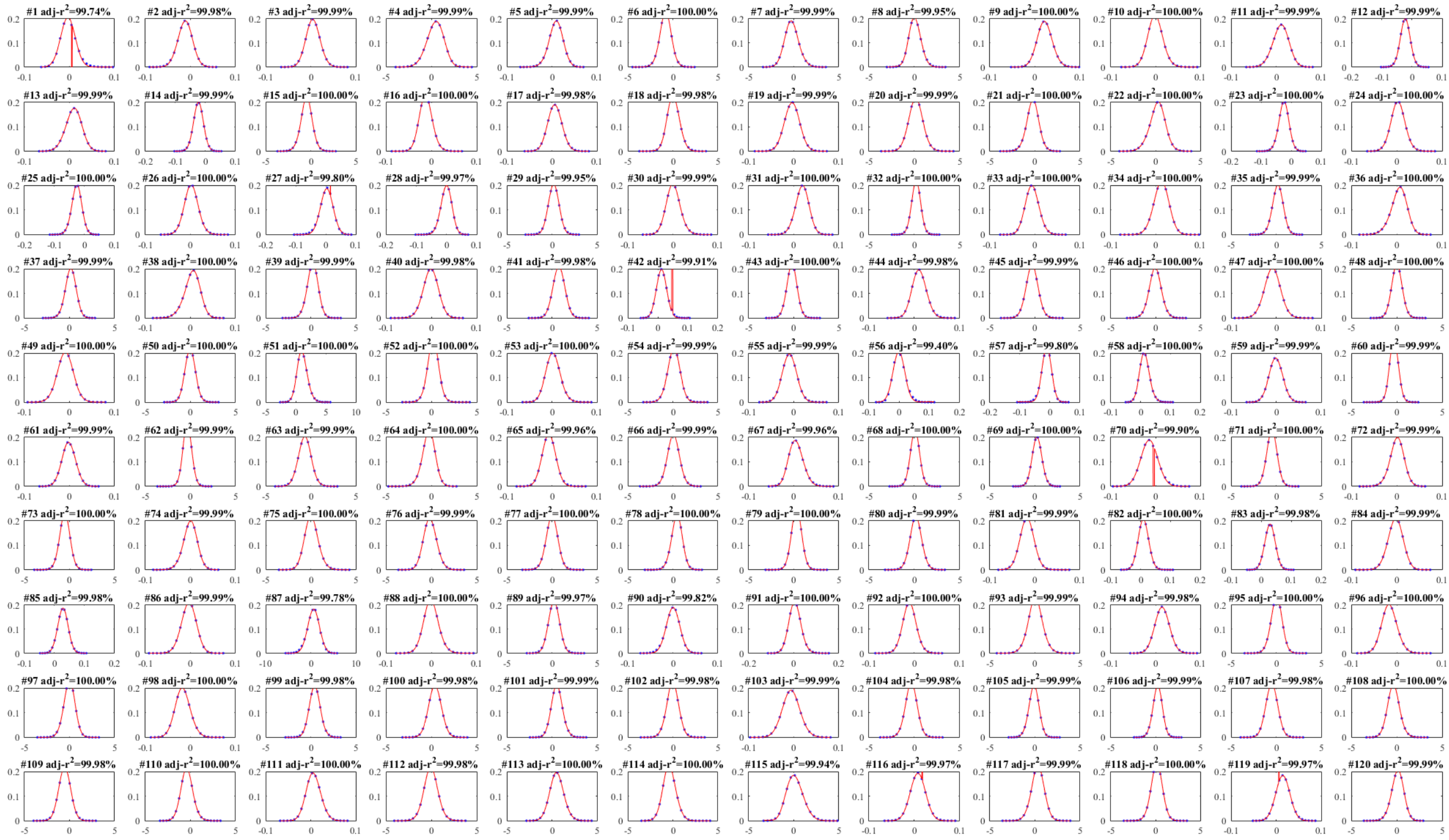}
  	\caption{Distributions of the first 120 channels in features generated by \modelname-$\mathcal{P}$ (the first 10 rows) and \modelname-$\mathcal{T}$ (the second 10 rows).}
  	\label{fig:gaussian}
\end{figure}

Distributions of the first 120 channels in features generated by \modelname-$\mathcal{P}$ and \modelname-$\mathcal{T}$ are plotted in Fig. \ref{fig:gaussian}. The number and $adj-R^2$ score of each channel can be found on top of each distribution. We can conclude that all variables follow Gaussian distribution.

In Fig. \ref{fig:correlation}, we visualize the correlation between different channels of the features (the first 256 from \modelname-$\mathcal{P}$ and the second 256 from \modelname-$\mathcal{T}$). As shown in the histogram, more than 99.3\% of the absolute values of coefficients are less than 0.3. We can conclude that the channels in the feature are independent from each other.

To sum up, the 512 dimensions in \modelname-$\mathcal{P}$ and \modelname-$\mathcal{T}$  both follow Gaussian distribution and are independent from each other, so the learned latent space is an ellipsoid and we can safely sample some points in it.

\section{More Examples Generated by \modelname{}}
\label{sec:more_examples}

To further prove the robustness and consistency of our model, in Fig. \ref{fig:moresample}, we generate more faces with various attributes changed. Taking \textit{Laugh} for example, we can observe: 

(1) For each identity (one row), the mouth changes from being closed (right) to being open (left). From right to left, while \textit{Laugh}-irrelevant attributes are well preserved, the opened mouth becomes wider and wider. Even the previously unseen teeth and tongue are generated with high fidelity. 

(2) Comparing different identities, the directions along which faces of different identities change are consistent (\eg, corners of the mouth moving upward, mouth opening wider). 

In conclusion, the \modelname{} framework is able to produce significant change for any target attribute, while keeping the change consistent across different identities.



\section{Full Comparison Results}
\label{sec:full_com_results}

\subsection{Attribute Classification Results}

Table \ref{tab: attribute} lists a complete ablation study of face attribute recognition results. The attribute IDs from 1 to 40 correspond to:
`5 o Clock Shadow', `Arched Eyebrows', `Attractive', `Bags Under Eyes', `Bald', `Bangs', `Big Lips', `Big Nose', `Black Hair', `Blond Hair', `Blurry', `Brown Hair', `Bushy Eyebrows', `Chubby', `Double Chin', `Eyeglasses', `Goatee', `Gray Hair', `Heavy Makeup', `High Cheekbones', `Male', `Mouth Slightly Open', `Mustache', `Narrow Eyes', `No Beard', `Oval Face', `Pale Skin', `Pointy Nose', `Receding Hairline', `Rosy Cheeks', `Sideburns', `Smiling', `Straight Hair', `Wavy Hair', `Wearing Earrings', `Wearing Hat', `Wearing Lipstick', `Wearing Necklace', `Wearing Necktie' and `Young'. It's easy to find that compared with \modelname-$\mathcal{P}$, \modelname-$\mathcal{T}$ wins on the most of ID-relevant attributes while the former one wins on the most of ID-irrelevant attributes.

\begin{table*}[h]
\centering
\caption{Complete ablation study on face attribute recognition.}
\begin{tabular}{c|c|c|c|c|c|c|c|c|c|c|c|c|c|c}
\hlinew{1.5pt}
Attribute ID                & 1    & 2    & 3    & 4    & 5    & 6    & 7    & 8    & 9    & 10   & 11   & 12   & 13   & 14   \\ \hline \hline
Baseline & 74.0 & 72.7 & 72.6 & 69.8 & 81.3 & 80.9 & 70.1 & 71.2 & 79.5 & 90.0 & 58.1 & 67.8 & 77.5 & 67.7 \\ 
\modelname-$\mathcal{P}$ w/o $\mathcal{L}_\mathcal{H}$ & 76.1 & 75.7 & 75.3 & 78.8 & 82.4 & 83.6 & 74.0 & 78.8 & 85.5 & 88.0 & 69.6 & 69.6 & 79.2 & 74.1 \\ 
\modelname-$\mathcal{P}$ w/o $\mathcal{L}_\mathcal{I}^\text{adv}$ & 75.9 & 74.2 & 73.4 & 76.0 & 82.3 & 80.8 & 71.7 & 74.0 & 83.3 & 90.2 & 62.5 & 69.8 & 76.9 & 72.1 \\ 
AutoEncoder+$\mathcal{L}_\mathcal{I}$ & 75.2 & 73.6 & 73.6 & 76.8 & 82.6 & 79.8 & 71.0 & 74.8 & 84.3 & 91.0 & 63.4 & 69.8 & 76.5 & 72.4 \\ 
AutoEncoder & 72.7 & 72.9 & 74.9 & 77.7 & 82.2 & 82.7 & 73.5 & 78.5 & 85.4 & 87.4 & 63.3 & 70.8 & 80.1 & 70.7 \\ \hline
\modelname-$\mathcal{T}$ & 78.7 & 76.6 & 76.7 & 76.1 & 84.1 & 82.9 & 74.2 & 76.4 & 84.8 & 93.9 & 67.3 & 71.8 & 82.1 & 73.9 \\ 
\modelname-$\mathcal{P}$ & 77.7 & 77.6 & 77.0 & 82.4 & 83.4 & 85.8 & 74.4 & 81.1 & 86.1 & 89.4 & 76.8 & 72.0 & 80.4 & 74.5 \\  \hline
\modelname{} & 78.8 & 78.1 & 79.2 & 83.1 & 84.8 & 86.5 & 75.2 & 81.3 & 87.4 & 94.2 & 78.4 & 72.9 & 83.0 & 74.6 \\ \hlinew{1.5pt}
\end{tabular}

\begin{tabular}{c|c|c|c|c|c|c|c|c|c|c|c|c|c|c|c|c}
\hlinew{1.5pt}
15   & 16   & 17   & 18   & 19   & 20   & 21   & 22   & 23   & 24   & 25   & 26   & 27   & 28   & 29   & 30   & 31   \\ \hline \hline
71.9 & 86.2 & 72.3 & 81.9 & 89.5 & 77.3 & 87.9 & 67.3 & 84.7 & 67.0 & 73.2 & 66.2 & 61.4 & 72.5 & 81.6 & 64.1 & 71.3 \\ 
78.4 & 83.5 & 77.5 & 85.5 & 88.5 & 85.8 & 89.0 & 79.0 & 82.7 & 77.0 & 74.3 & 76.3 & 87.0 & 78.0 & 85.3 & 76.8 & 78.5 \\ 
75.8 & 84.7 & 71.6 & 82.4 & 90.2 & 84.3 & 90.6 & 73.9 & 82.9 & 72.9 & 73.1 & 71.6 & 80.6 & 75.0 & 82.7 & 68.8 & 76.2 \\ 
75.5 & 85.0 & 72.4 & 82.2 & 91.0 & 83.5 & 90.6 & 74.8 & 82.1 & 73.1 & 72.2 & 72.0 & 79.9 & 75.4 & 83.2 & 68.0 & 75.3 \\ 
75.7 & 75.5 & 78.0 & 82.3 & 88.1 & 85.4 & 85.2 & 74.6 & 82.1 & 68.2 & 74.0 & 75.6 & 89.8 & 75.8 & 84.2 & 61.9 & 76.9 \\ \hline
76.9 & 88.5 & 76.4 & 85.7 & 94.0 & 83.4 & 94.0 & 75.2 & 86.9 & 72.6 & 76.5 & 72.5 & 72.3 & 76.7 & 85.6 & 68.2 & 77.6 \\ 
79.9 & 86.7 & 78.6 & 86.0 & 92.5 & 88.2 & 92.3 & 81.1 & 84.1 & 77.5 & 76.3 & 77.6 & 89.4 & 78.8 & 87.3 & 79.3 & 80.3 \\  \hline
80.2 & 89.5 & 78.6 & 86.9 & 94.5 & 88.8 & 94.3 & 81.7 & 87.3 & 77.5 & 77.7 & 78.7 & 89.8 & 79.8 & 88.0 & 79.9 & 80.5 \\  \hlinew{1.5pt}
\end{tabular}

\begin{tabular}{c|c|c|c|c|c|c|c|c|c}
\hlinew{1.5pt}
32   & 33   & 34   & 35   & 36   & 37   & 38   & 39   & 40   & Ave.  \\ \hline \hline
79.2 & 67.2 & 70.0 & 87.7 & 73.4 & 88.3 & 79.5 & 75.4 & 76.6 & 75.2 \\ 
89.4 & 73.8 & 82.0 & 89.4 & 77.9 & 88.7 & 82.7 & 77.3 & 79.7 & 80.4 \\ 
85.9 & 71.4 & 77.2 & 89.7 & 75.7 & 91.2 & 81.7 & 78.1 & 78.4 & 78.2 \\ 
85.3 & 70.7 & 77.5 & 89.9 & 76.3 & 90.4 & 82.5 & 77.3 & 77.9 & 78.2 \\
89.1 & 73.1 & 82.2 & 87.7 & 78.7 & 87.1 & 83.2 & 75.5 & 76.3 & 78.5 \\  \hline
85.2 & 69.1 & 75.0 & 89.7 & 75.5 & 90.5 & 82.3 & 78.2 & 78.3 & 79.7 \\ 
91.8 & 73.0 & 80.4 & 89.2 & 79.4 & 89.0 & 83.1 & 77.4 & 79.0 & 81.9 \\  \hline
92.2 & 73.6 & 81.7 & 89.7 & 80.5 & 91.4 & 84.0 & 78.7 & 79.2 & \textbf{83.1} \\  \hlinew{1.5pt}
\end{tabular}
\label{tab: attribute}
\end{table*}

\subsection{Identity Preserving Results}

In Table \ref{tab:id}, we present a comparison of results on face  verification on the LFW dataset.

\begin{table}[h!]
\centering
\caption{Comparison of results on face verification.}
\label{tab:id}
\begin{tabular}{c|c|c}
\hlinew{1.5pt}
Methods                 & Mean Acc. & TPR@0.001FPR \\ \hline  \hline 
Basel.+WebFace          & 98.93     & 94.87        \\ 
\modelname +WebFace     & \underline{99.25}      & \underline{96.80}         \\ \hline
Basel.+MS1M   			& \textbf{99.816}     & \textbf{99.73}        \\ 
\modelname-$\mathcal{T}$+MS1M          & 99.78     & 99.63        \\ 
\modelname{}+MS1M    & \underline{99.80}     & \underline{99.4}         \\ \hline \hline
THU CV-AI Lab        & 99.73     & 99.2         \\ \hline
DeepID3 \cite{SunLWT15}& 99.53     & 97.4         \\ \hline
Faceall              & 99.67     & 97.1         \\ \hline
DeepID2+ \cite{sun2015}& 99.47     & 96.7         \\ \hline
DeepID2 \cite{sun2014c}& 99.15     & 96.33        \\ \hline
DLIB                 & 99.38     & 95.87        \\ \hline
GaussianFace \cite{lu2014} & 98.52     & 93.9         \\ \hline
DeepID \cite{sun2014a}& 97.45     & 87.57        \\ \hline
TL Joint Bayesian \cite{cao2013}& 96.33     & 83.67        \\ \hline
Human+cropped \cite{kumar2009}& 97.53     & 82.63        \\ \hline
DeepFace \cite{parkhi2015}& 97.35     & 82.17        \\ \hline
high dimensional LBP \cite{chen2013}& 95.17     & 76.43        \\ 
\hlinew{1.5pt}
\end{tabular}
\end{table}

{\small
\bibliographystyle{ieee}
\bibliography{gan_bib}
}

\end{document}